\documentclass[10pt,twocolumn,letterpaper]{article}

\usepackage[pagenumbers]{cvpr} % To force page numbers, e.g. for an arXiv version

\definecolor{cvprblue}{rgb}{0.21,0.49,0.74}
\usepackage[pagebackref,breaklinks,colorlinks,allcolors=cvprblue]{hyperref}
\usepackage{multirow}
\usepackage{arydshln}
\usepackage{float}
\usepackage{tablefootnote}
\usepackage{tikz}

\usepackage{lineno}
\title{One-D-Piece: Image Tokenizer Meets Quality-Controllable Compression}

\author{
    Keita Miwa \quad Kento Sasaki \quad Hidehisa Arai \quad Tsubasa Takahashi \quad Yu Yamaguchi \\
    Turing Inc. \\
    {\tt\small \url{https://turingmotors.github.io/one-d-piece-tokenizer}}
}

\begin{document}
\twocolumn[{
\renewcommand\twocolumn[1][]{#1}
\maketitle
\begin{center}
    \centering
    \includegraphics[width=0.99\linewidth]{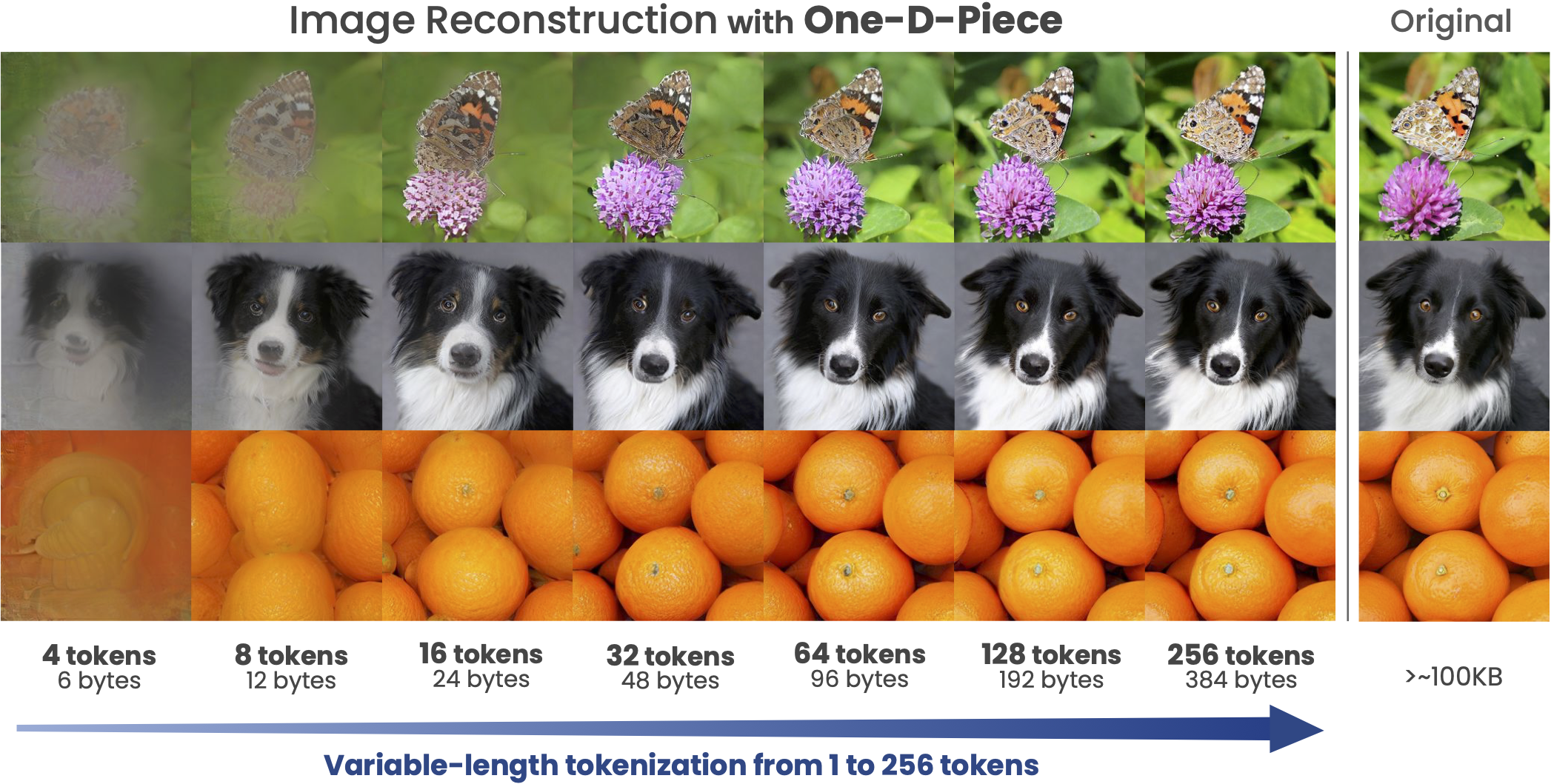}
    \captionof{figure}{We propose \textbf{One-D-Piece}, discrete image tokenizer that enables variable-length tokenization adjustable from 1 to 256 tokens. Even with a very small number of tokens (e.g., \( n_{\text{tokens}}=8 \)), it achieves recognizable image reconstructions. As the token count increases, the image quality progressively improves, reaching near-original fidelity at \( n_{\text{tokens}}=256 \).
    \label{fig:overview}
    }
\end{center}}]

\begin{abstract}
Current image tokenization methods require a large number of tokens to capture the information contained within images. Although the amount of information varies across images, most image tokenizers only support fixed-length tokenization, leading to inefficiency in token allocation.
In this study, we introduce One-D-Piece, a discrete image tokenizer designed for variable-length tokenization, achieving quality-controllable mechanism. To enable variable compression rate, we introduce a simple but effective regularization mechanism named ``Tail Token Drop'' into discrete one-dimensional image tokenizers. This method encourages critical information to concentrate at the head of the token sequence, enabling support of variadic tokenization, while preserving state-of-the-art reconstruction quality.
We evaluate our tokenizer across multiple reconstruction quality metrics and find that it delivers significantly better perceptual quality than existing quality-controllable compression methods, including JPEG and WebP, at smaller byte sizes. Furthermore, we assess our tokenizer on various downstream computer vision tasks, including image classification, object detection, semantic segmentation, and depth estimation, confirming its adaptability to numerous applications compared to other variable-rate methods. Our approach demonstrates the versatility of variable-length discrete image tokenization, establishing a new paradigm in both compression efficiency and reconstruction performance.
Finally, we validate the effectiveness of tail token drop via detailed analysis of tokenizers.
\textbf{Codes will be available after the acceptance of the paper.}
\end{abstract}
    
\section{Introduction}
\label{sec:intro}

In recent years, with the rapid advancements in vision-language models (VLMs)~\cite{liu2024llavanext, Qwen-VL, laurencon2024what} and image and video generation models~\cite{yan2021videogpt, villegas2022phenaki, bruce2024genie}, the concept of discrete tokenization for visual data, similar to language tokenization, has garnered increasing attention~\cite{NIPS2017_7a98af17_vqvae, yu2024language_magvitv2, luo2024open_openmagvitv2, wang2024omnitokenizer}. This approach enables seamless integration with Transformer-based models~\cite{vaswani2017_attention}, simplifying model architectures and reducing computational complexity.

However, challenges remain in discrete tokenization for visual data, particularly in capturing spatial structures, which often requires long, fixed-length token sequences. For instance, typical image tokenizers require as much as 256 tokens to represent a single 256×256 pixel image, which limits their flexibility in practical applications. To overcome this limitation, one-dimensional (1D) tokenization methods have emerged, aiming to achieve higher compression rates while maintaining reconstruction quality. Notably, the SEED tokenizer~\cite{ge2024making_seed_tokenizer} can semantically represent images by causal 1D sequences, while the TiTok tokenizer~\cite{yu2024an_titok} achieves high-quality image reconstruction with just 32 tokens. These approaches efficiently encode the entire image into a compact 1D sequence, significantly reducing the number of tokens required.
Nevertheless, there exists a fundamental trade-off between compression rate and reconstruction quality~\cite{shannon1959coding, blau2019rethinking}; higher compression results in greater degradation, especially for complex images. As current image tokenizers are designed to produce fixed number of tokens, it is impossible to control the quality based on specific requirements.

In contrast, classical image compression methods such as JPEG~\cite{wallace1991jpeg} have long addressed this trade-off by allowing users to adjust compression rates based on the desired quality. These methods provide well-established mechanisms for balancing file size and visual fidelity, making them highly versatile across various applications. 
However, traditional compression algorithms are not designed for direct use as input representations for neural networks, making it challenging to integrate them into neural models like VLMs. Furthermore, these algorithms differ fundamentally from the adaptive, model-driven strategies used in modern image tokenizers, complicating the transfer of established compression techniques to the tokenization domain.

\begin{figure}[bt]
    \centering
    \includegraphics[width=\columnwidth]{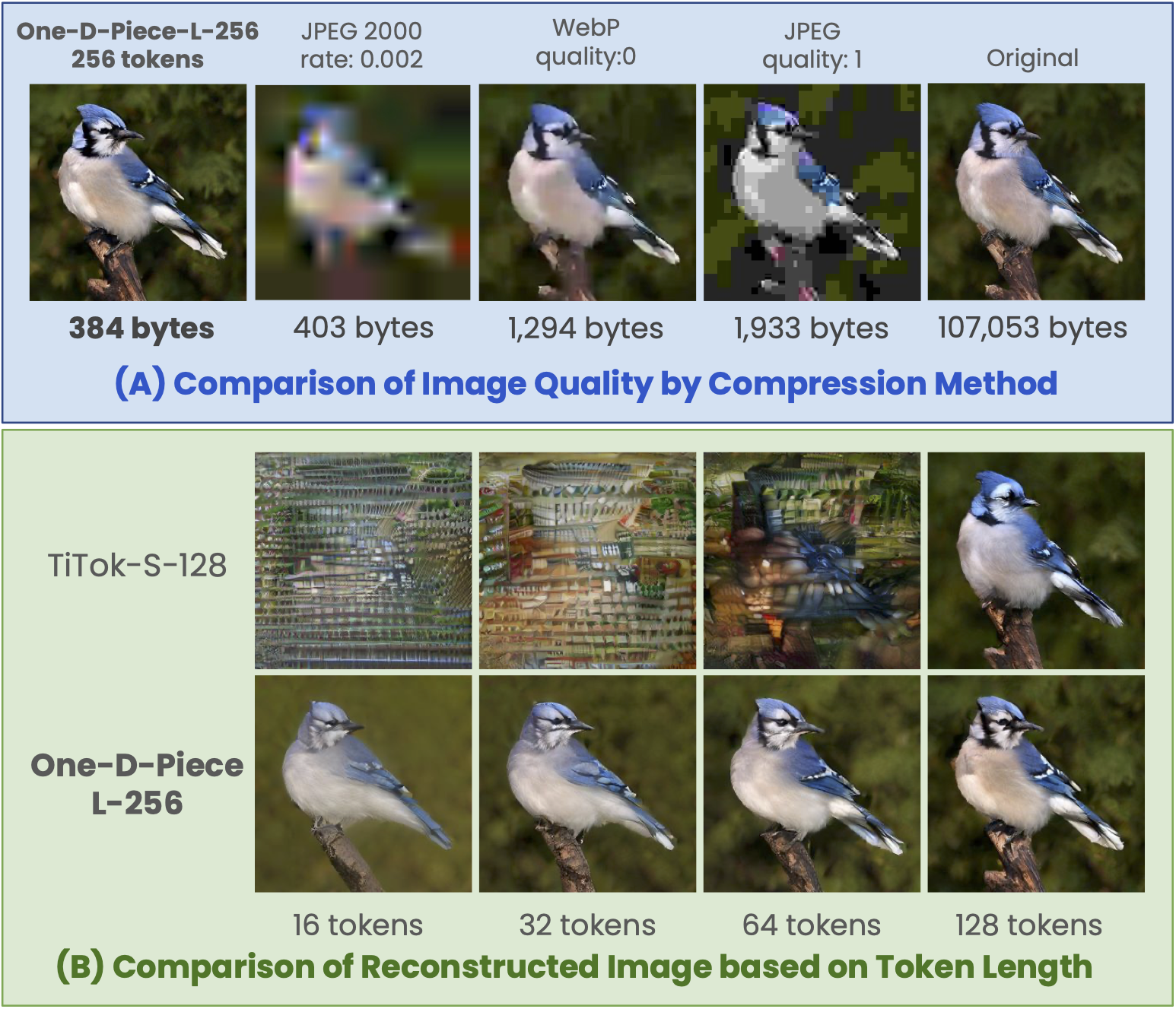}
    \caption{
    \textbf{Comparison of Image Quality and Compression Efficiency.} 
    (A) One-D-Piece-L-256 achieves superior visual quality with enhanced compression efficiency, reducing image size to 384 bytes, outperforming image formats. 
    (B) One-D-Piece-L-256 exhibits clear improvements in reconstruction quality as the token length increases, demonstrating effectiveness even with fewer tokens (e.g., 16, 32, 64, and 128 tokens). while TiTok-S-128 struggles with shorter token lengths due to fixed token limitations.
    }
    \label{fig:comparison_quality_efficiency}
\end{figure}

Given these differences, there is a pressing need for a novel approach that bridges the adaptability of tokenization methods with the efficiency of traditional compression formats. To address this challenge, we propose \textit{One-D-Piece}, a new variable-length discrete image tokenizer that combines the benefits of tokenization with the flexibility of classical compression methods. Our approach introduces a simple yet effective regularization technique, \textit{Tail Token Drop}, which concentrates critical information at the beginning of the token sequence, enabling efficient and adaptive token lengths ranging from 1 to 256 tokens. This allows for maintaining high reconstruction quality even with a low token count, providing flexible compression suited for diverse applications. As illustrated in \autoref{fig:overview} and \autoref{fig:comparison_quality_efficiency}, our model supports variable-length image tokenization and can effectively produce visually accurate tokenizations and reconstructions even with as few as 8 or 16 tokens.

We further evaluate our method not only using standard image quality metrics but also by assessing its performance on a range of downstream tasks, demonstrating practical benefits over classical image compression techniques. We showcase the effectiveness of One-D-Piece through extensive experiments on various computer vision tasks, including image classification, object detection, and semantic segmentation. Our results indicate that One-D-Piece outperforms existing variable-length compression methods, including JPEG~\cite{wallace1991jpeg}, JPEG 2000~\cite{iso_jpeg2000_2019}, and WebP~\cite{webp}, in terms of perceptual quality, especially at low token counts.

Our main contributions can be summarized as follows:
\begin{itemize}
    \item We introduce One-D-Piece, a variable-length discrete image tokenizer utilizing a novel Tail Token Drop regularization technique.
    \item Our method achieves competitive reconstruction quality while supporting flexible token lengths, making it a versatile solution for diverse compression needs.
    \item We validate our approach through comprehensive evaluations on both perceptual quality and downstream task performance, demonstrating its superiority over traditional image compression methods.
    \item We analyze the behaviors of our models in detail, revealing our Tail Token Drop method effectively works to concentrating important information at the head.
\end{itemize}

\section{Related Work}
\label{sec:realated_work}

Recent advancements in computer vision and machine learning have increasingly relied on efficient data representations to improve model performance and scalability. One critical area of focus is image tokenization, which aims to transform high-dimensional image data into more compact and efficient forms.

\subsection{Image Tokenization}

Pixel-space image representations are highly redundant, and a variety of AutoEncoder-based approaches, such as VAE~\cite{Kingma2014_vae}, are widely studied as methods to extract essential information into lower-dimensional latent spaces. Discrete image tokenization methods, like those based on VQ-VAE~\cite{NIPS2017_7a98af17_vqvae}, play a crucial role in tasks such as compression, image and video generation~\cite{esser2021taming, chang2022maskgit, weber2024maskbit, yan2021videogpt, villegas2022phenaki, bruce2024genie}, as well as in VLMs~\cite{chameleonteam2024chameleonmixedmodalearlyfusionfoundation, wang2024emu3, ge2024seed}. Particularly in transformer-based generative models, discrete image tokenizers are employed to decompose images into sequence of tokens, enabling efficient representation and manipulation.

Typical approaches in discrete image tokenization utilize Convolutional Neural Networks (CNNs) within an AutoEncoder framework, incorporating vector quantization, to capture local information effectively~\cite{NIPS2017_7a98af17_vqvae, yu2024language_magvitv2, luo2024open_openmagvitv2, wang2024omnitokenizer, esser2021taming, weber2024maskbit, chang2022maskgit}. These methods are classified as 2D tokenizers, as they maintain a two-dimensional structural representation of the image. While CNN-based 2D tokenizers are straightforward for aggregating local features, they struggle to integrate global information, often resulting in low compression efficiency. Typical 2D tokenizers consume 256 tokens for 256$\times$256 images.

Recent advances have introduced 1D tokenizers that compress images by capturing local and global information simultaneously, using architectures like Transformers to process the entire image context. For instance, SEED-Tokenizer~\cite{ge2024making_seed_tokenizer} uses Vision Transformer (ViT)~\cite{dosovitskiy2021vit} and reduces an image to a sequence of semantic tokens, demonstrating efficient compression for use in VLMs~\cite{ge2024seed, wang2024emu3}.

Another high-compression 1D tokenizer, TiTok~\cite{yu2024an_titok}, effectively compresses images into just 32 sequential tokens while preserving high visual fidelity, outperforming existing 2D tokenizers in both compression efficiency and reconstruction quality. However, it is well-studied within rate-distortion theory that high compression is not a free-lunch~\cite{shannon1959coding, blau2019rethinking}. There is a trade-off between reconstruction quality and compression rate, and this trade-off becomes severe for complex images, while it is not possible to adjust reconstruction quality with current image tokenizers.

\subsection{Variable-Length Coding}

The issue of rate-distortion trade-offs is partially addressed by quality control mechanisms in standardized image compression algorithms. Variable-rate compression enables usage-adaptive byte size control and enhances flexibility across diverse applications.

Standard image compression methods like JPEG~\cite{wallace1991jpeg}, JPEG 2000~\cite{iso_jpeg2000_2019}, and WebP~\cite{webp} utilize advanced techniques for efficient variable-rate compression. JPEG employs the Discrete Cosine Transform (DCT) for frequency domain transformation, followed by quantization and Huffman coding, enabling flexible compression rates. JPEG 2000 improves upon this by using Discrete Wavelet Transform and embedded block coding, providing compression controllability. 
WebP, building on techniques from VP8 video compression, uses block prediction and DCT-based coding, achieving better compression ratios than JPEG at comparable quality levels.

Adaptive quality control has also been explored in the field of neural image compression. ``Tail Drop''~\cite{koike_wang_2020_taildrop} is a technique that concentrates critical information towards \textit{head} part of the embeddings. This method is applied in AutoEncoder-based models by imposing a higher dropout rate on the \textit{tail} of the latent representation. By focusing essential information towards the beginning of the latent sequence, tail drop enhances compression efficiency, allowing for variable-rate compression.
Tail drop is simpler and more suited for neural discrete image tokenizers than algorithms-based image compression methods. However, it faces a key limitation: 2D tokenizers generate spatially structured tokens, each tied to specific image regions. This spatial dependency complicates the application of tail drop without disrupting spatial alignment.

\section{Method}

We introduce \textbf{One}-\textbf{D}imensional Image \textbf{Piece} Tokenizer (One-D-Piece), a novel discrete tokenizer designed for efficient image compression. 
In contrast to existing image tokenizers that typically generate fixed-length sequences, One-D-Piece produces variable-length tokens, similar to text tokenizers like WordPiece~\cite{wu2016google} and SentencePiece~\cite{kudo2018sentencepiecesimplelanguageindependent}.

\subsection{Tail Token Drop}
\begin{figure}[ht]
    \centering
    \includegraphics[width=\columnwidth]{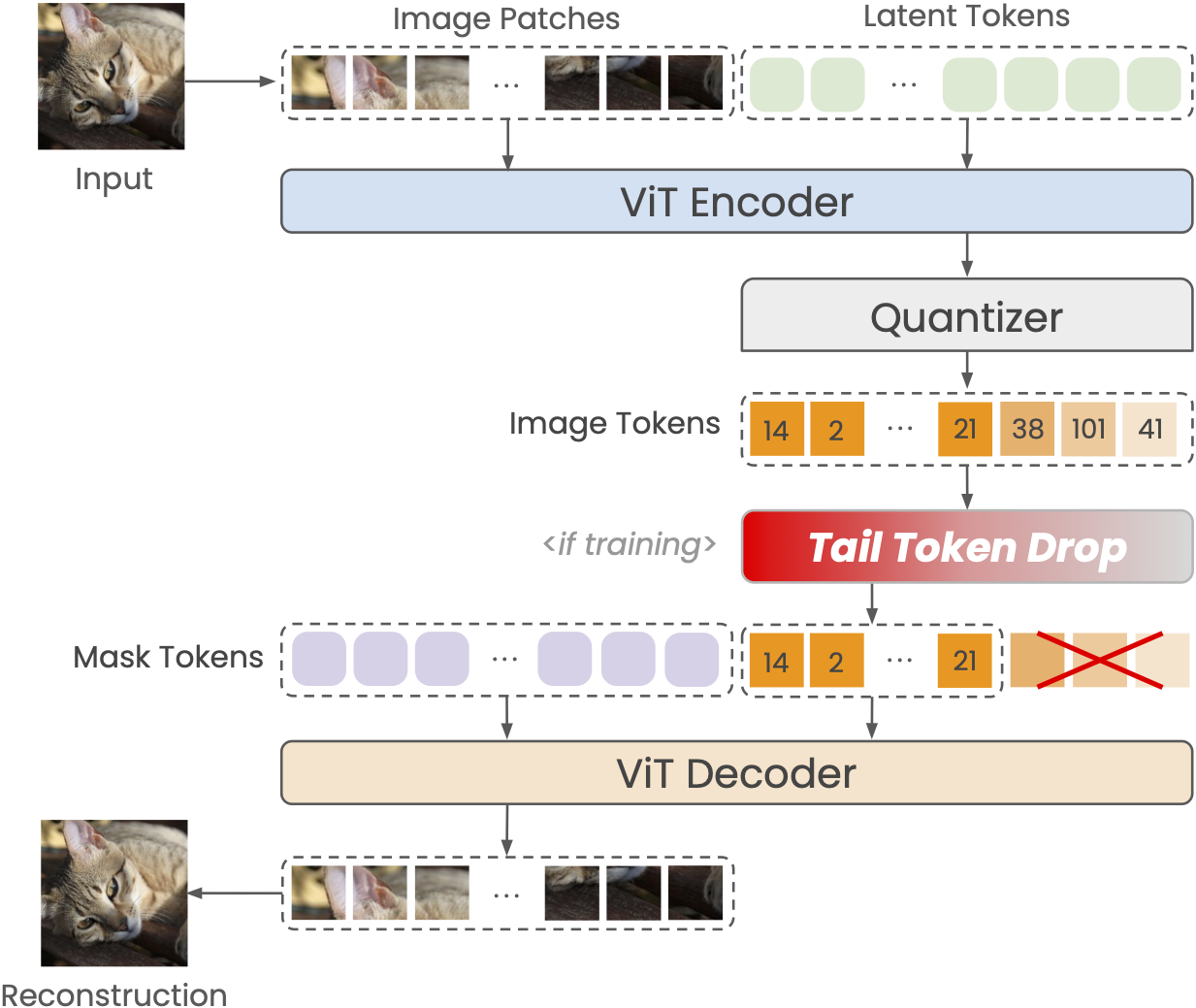}
    \caption{One-D-Piece applies random Tail Token Drop during training to concentrate the most important information at the head of the token sequence.}
    \label{fig:architecture}
\end{figure}

A regularization technique called ``tail drop'' was originally introduced to improve compression efficiency by dynamically controlling the dimensionality of the latent representation~\cite{koike_wang_2020_taildrop}. Tail drop can be viewed as a variant of Dropout, where the dropout rate progressively increases towards the end (tail) of the latent vector. This technique prioritizes learning the most essential features in the earlier (head) neurons, while gradually discarding less critical information in the tail. As a result, tail drop enables flexible compression, where the dimensionality can be adjusted at inference time with controlling the quality of the reconstruction.

We adapt this simple yet effective method for image tokenization with modifications, introducing ``Tail Token Drop,'' which involves randomly truncating the tail of the token sequence and can be applied to 1D tokenizers, which produces structure-free image tokens.

Formally, let \( \mathbf{q} = [q_1, q_2, \dots, q_N] \) be the token sequence generated by our 1D tokenizer, where \( q_i \) represents each token in the sequence, and \( N \) is the total number of tokens. 
During training, the number of tokens to be dropped, \( k \), is sampled from a uniform distribution:

\[
k \sim U(0, N-1)
\]

Thus, the token sequence after applying Tail Token Drop, denoted as \( \mathbf{q'} \), is given by:

\[
\mathbf{q'} = [q_1, q_2, \dots, q_{N - k}]
\]

By applying this regularization, the tokenizer is encouraged to accumulate more critical information towards the beginning of the sequence by randomly truncating tokens at the tail end and less significant information tends to accumulate at the end, which is more likely to be truncated. As a result, the tokenizer trained with Tail Token Drop technique allows the token sequence length to be flexibly adjusted based on the information content of the image, by cutting the tails adaptively.

\subsection{Architecture}

The architecture of One-D-Piece focuses on two important requirements.
First, the tokens produced must form a 1D sequence. They should not explicitly correspond to 2D structure, like most 2D tokenizers. This restriction exists because the tail cannot be defined for 2D tokens, hindering application of Tail Token Drop.
Second, the detokenizer must handle variable-length of tokens; otherwise, the Tail Token Drop technique cannot be applied. 

To meet these requirements, we build upon the TiTok architecture for One-D-Piece.
The TiTok model~\cite{yu2024an_titok} consists of three main components: the encoder, quantizer, and decoder. 
The encoder first divides the input image \( \mathbf{X} \in \mathbb{R}^{H \times W \times C} \) patches. Each patch is embedded into a vector, yielding a set of patch embeddings. We also include \(N\) learnable latent tokens. The output of the ViT encoder, $\mathbf{z}$, corresponding to the latent tokens is used as latent space. 
The output \( \mathbf{z} \) is then discretized by the quantizer, producing a discrete set of token representations \( \mathbf{q} = \text{Quantizer}(\mathbf{z}) \). This quantized sequence serves as a discrete representation of the image. For One-D-Piece training, we apply Tail Token Drop regularization to this as $ \mathbf{q}' = \text{TailTokenDrop}(\mathbf{q})$. 
In the decoder, the quantized tokens \( \mathbf{q} \) or \( \mathbf{q}' \) are processed alongside mask tokens. The output corresponding to the mask tokens is then upscaled by a CNN to generate the final reconstructed image \( \hat{\mathbf{X}} \in \mathbb{R}^{H \times W \times C} \).

\begin{table*}[t]
\setlength{\dashlinedash}{0.5pt}
\setlength{\dashlinegap}{1.5pt}
\setlength{\arrayrulewidth}{0.5pt}
\centering
\begin{tabular}{c|c|c|c|c|c}
\hline
\textbf{Method}                              & \textbf{Mechanism}                     & \textbf{Token Counts} & \textbf{Byte per Image} & \textbf{rFID}$\downarrow$ & \textbf{PSNR}$\uparrow$ \\ \hline
VQGAN~\cite{esser2021taming}                 &                                       & 256       & 320                    & 7.94          & 19.4    \\
MaskGIT~\cite{chang2022maskgit}              & \multirow{2}{*}{CNN-based}            & 256       & 320                    & 2.28          & ---     \\
LlamaGen~\cite{sun2024autoregressive}        & \multirow{2}{*}{2D Discrete Tokenizer}& 256       & 448                    & 2.19          & 20.79   \\
VQGAN+~\cite{weber2024maskbit}               &                                       & 256       & 384                    & 1.61          & ---     \\
Open-MAGVIT2~\cite{luo2024open_openmagvitv2} &                                       & 256       & 576                    & 1.17          & 21.90   \\ \hdashline
JPEG~\cite{wallace1991jpeg}                  &  \multirow{2}{*}{Algorithm-based}     & ---       & 2,063.3                & 113.3         & 20.99   \\
JPEG 2000~\cite{iso_jpeg2000_2019}           &  \multirow{2}{*}{Image Formats}       & ---       & 406.3                  & 299.4         & 19.19   \\ 
WebP~\cite{webp}                             &                                       & ---       & 1,964.0                & 31.98         & \textbf{26.18}   \\ \hdashline
TiTok-S-128                                  & \multirow{2}{*}{ViT-based}            & 128       & 192                    & 1.70          & 17.80  \\
TiTok-B-64                                   & \multirow{2}{*}{1D Discrete Tokenizer}& 64        & 96                     & 1.71          & 17.13   \\
TiTok-L-32                                   &                                       & 32        & 48                     & 2.21          & 15.96   \\ \hline
One-D-Piece-S-256                            & ViT-based                             & 1 to 256  & 1.5 to 384              & 1.48          & 18.28   \\
One-D-Piece-B-256                            & 1D Discrete Tokenizer                 & 1 to 256  & 1.5 to 384              & \textbf{1.11} & 18.77   \\
One-D-Piece-L-256                            & with Tail Token Drop                  & 1 to 256  & 1.5 to 384              & \textbf{1.08} & 19.04   \\ \hline

\end{tabular}
\caption{\textbf{Reconstruction Quality across Tokenizers and Image Formats}. Other than algorithm-based image formats, One-D-Piece is the only model which support variable-length image tokenization. For these formats, byte per Image values represent can vary based on image content and quality settings.}
\label{tab:reconstruction_quality_rfid_psnr}
\end{table*}

\subsection{Training}

We use the two-stage training strategy adopted by TiTok. In the first stage, the model is trained to predict the logits of a pretrained tokenizer using cross-entropy loss. 
The second stage involves training the model to reconstruct the image itself after learning the logits in the first stage. Here, the pretrained tokenizer is incorporated into the decoder, while the encoder remains frozen. The model is then optimized using reconstruction loss. The loss function includes L2 loss to reduce distortion, and perceptual loss and GAN loss for improved visual quality.
To support variable-length tokenization, we apply Tail Token Drop during training to dynamically adjust token lengths.
For each batch, an index from 1 to 256 (representing minimum to maximum) is uniformly sampled, and tokens beyond this index are truncated.

\[
\mathcal{L}_{\textrm{stage2}} = \mathcal{L}_{\textrm{L2}} + \mathcal{L}_{\textrm{Perceptual}} + \mathcal{L}_{\textrm{GAN}}
\]

Training and evaluation are both conducted using the ImageNet-1K dataset~\cite{deng2009imagenet}, which contains 1,000 object classes, 1,281,167 training images, 50,000 validation images, and 100,000 test images.

\section{Experiments}
\label{sec:experiments}

The primary goal of these experiments is to evaluate the effectiveness of our One-D-Piece across different settings, including reconstruction quality, as well as various downstream tasks. By comparing against compression algorithms including JPEG, JPEG 2000, and WebP and existing image tokenizers, we demonstrate the advantages of our approach.

We train One-D-Piece models using a maximum of 256 tokens per image. To explore different model complexities, we trained three variants, S-256, B-256, and L-256, which differ in the parameter size of the Vision Transformer. The hyperparameter settings strictly followed those used for TiTok.

Training was conducted in two phases: 100 epochs on ImageNet-1K in Stage 1, followed by 200 epochs in Stage 2. We utilize 8 NVIDIA H100  80 GB GPU. Training times vary based on model complexity: S-256 takes approximately 5 days, B-256 requires about 6 days, and L-256 takes around 9 days. This difference reflects the increased parameter count and computational demands of larger models.

\subsection{Reconstruction}

\begin{figure*}[htbp]
    \centering
    \begin{minipage}{0.37\linewidth}
        \centering
        \includegraphics[width=\linewidth]{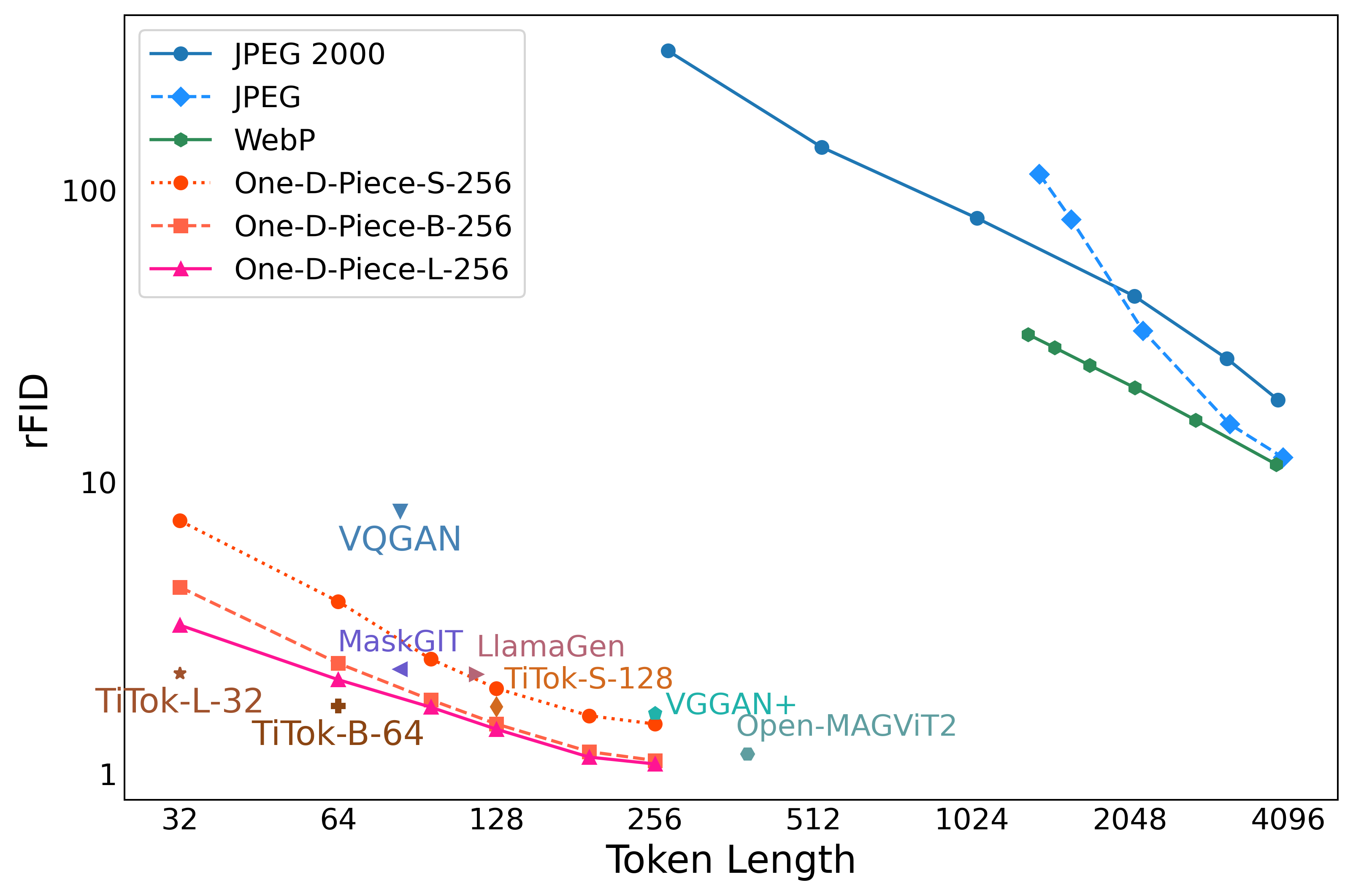}
        \caption{\textbf{Comparison of rFID by Token Length}. One-D-Piece improves rFID as token length increases and achieves better scores with fewer tokens than standard image formats (token counts for standard formats are converted from bytes per image).}
        \label{fig:length_to_rfid}
    \end{minipage}
    \hfill
    \begin{minipage}{0.61\linewidth}
        \centering
        \includegraphics[width=\linewidth]{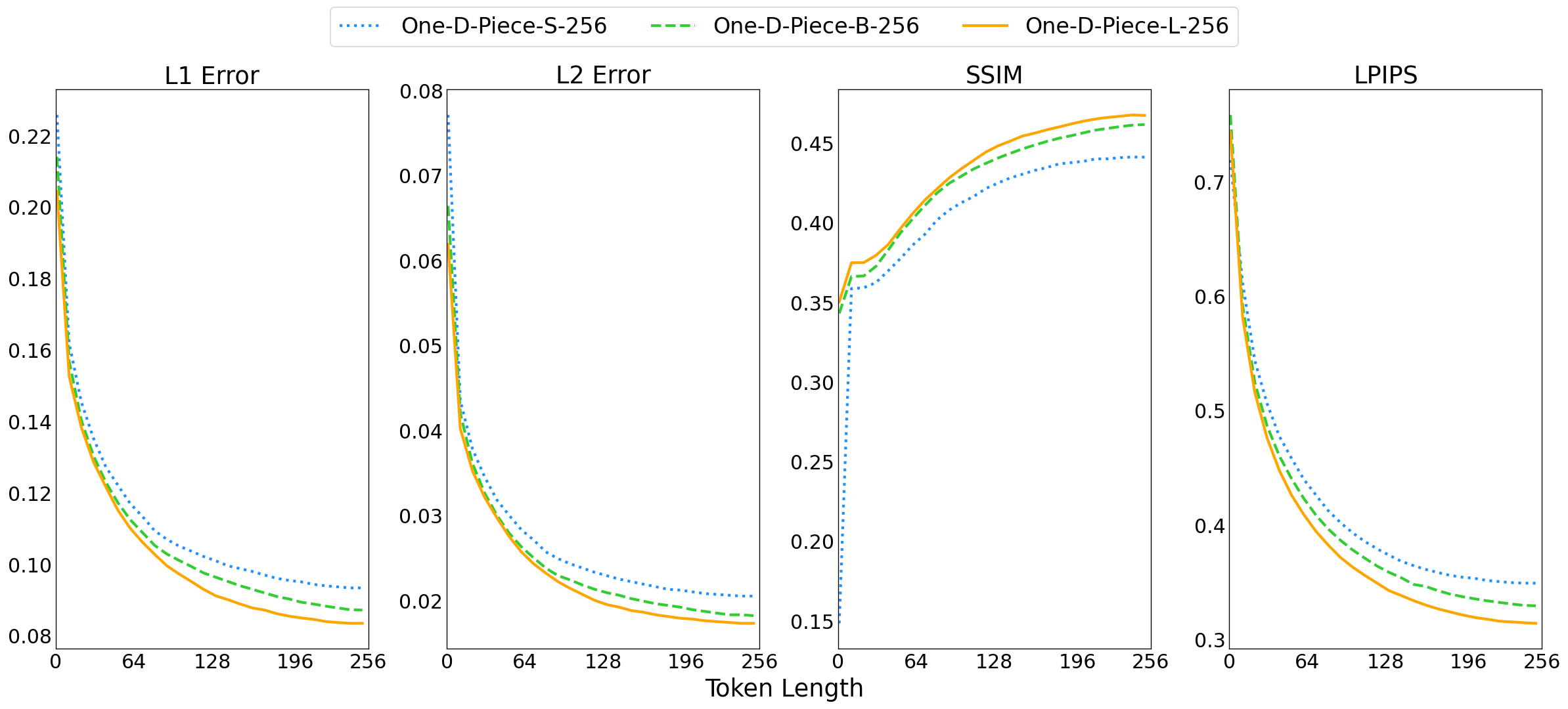}
        \caption{\textbf{Reconstruction Quality by Token Length}. L1 Error(\(\downarrow\)), L2 Error(\(\downarrow\)), SSIM(\(\uparrow\)), and LPIPS(\(\downarrow\)) for token lengths from 1 to 256. Reconstruction quality improves steadily with increasing token length. Larger models achieve faster gains with fewer tokens and higher performance with more tokens.}
        \label{fig:reconstruction_quality}
    \end{minipage}
\end{figure*}

We evaluate the reconstruction quality of three variants of the One-D-Piece models and compare them to other image tokenizers and standard image formats.

\autoref{tab:reconstruction_quality_rfid_psnr} includes our main metrics, Fréchet Inception Distance (FID)~\cite{heusel2017gans} for assessing perceptual quality, and PSNR for measuring distortion. Our models exhibit sufficiently better performance compared to other image tokenizers, with the B-256 and L-256 variants achieving the lowest rFID scores of 1.11 and 1.08 at 256 tokens.

For image formats, we controlled the quality settings to align the average bytes per image with our models. 
Specifically, we converted the token count to bytes using a rate of 1.5 bytes per token (12 bits). 
For JPEG and WebP, we used the minimum quality settings to minimize byte size. 
For JPEG 2000, we verified that a target compression rate of 0.002 yielded an average byte size of 406.3, comparable to our models with 256 tokens. 
Notably, these traditional image formats performed poorly on the rFID metric compared to neural image tokenizers, including our models.

It is worth mentioning that the PSNR metric shows opposite trends; our models exhibit relatively lower performance compared to image formats. This is due to training objective of our models, which includes GAN loss and perceptual loss, focusing on improving perceptual quality rather than reducing pixel-level distortion. This result aligns with~\cite{blau2019rethinking}, where rate-distortion-perception trade-off is reported.
Nevertheless, as we can confirm in \autoref{fig:comparison_quality_efficiency}, our tokenizers exhibit far better perceptual quality compared to these image formats with supporting variable-length tokenization.

To examine the reconstruction quality as the token length varies, we provide the rFID plot in \autoref{fig:length_to_rfid}. The rFID steadily improves as more tokens become available, and our models exhibit far better efficiency compared to algorithm-based image compression formats. We also compare the rFID of our models with TiTok variants in \autoref{tab:titok_onedpiece_comparison}. While our models exhibit the lowest rFID with 256 tokens, our models do not match TiTok at token counts of 32, 64, and 128 for equivalent model sizes. We attribute this result to the introduction of the Tail Token Drop. This highlights a challenge for improving the perceptual quality at smaller token size, but we leave it for the future exploration.

We further illustrate the behavior of L1 error, L2 error, structural similarity (SSIM), and LPIPS~\cite{zhang2018perceptual} with VGG~\cite{simonyan2014very}, along with the token length in the \autoref{fig:reconstruction_quality}. The quality continues to improve, demonstrating the effectiveness of our Tail Token Drop approach.

\begin{table}[bp]
\resizebox{\linewidth}{!}{
\centering
\begin{tabular}{ccccc}
\hline
\multirow{2}{*}{\textbf{Model}} & \multicolumn{3}{c}{\textbf{rFID}$\downarrow$} & \multirow{2}{*}{\textbf{LPA}$\uparrow$} \\ \cmidrule(lr){2-4}
& \textbf{@32} & \textbf{@64} & \textbf{@128} &   \\ \hline
\textbf{TiTok-S-128}       &  --- & ---  & 1.70 & 0.349           \\
\textbf{TiTok-B-64}        &  --- & 1.71 &  --- & 0.276  \\
\textbf{TiTok-L-32}        & 2.21 & ---  &  --- & 0.281 \\ \hline
\textbf{One-D-Piece-S-256} & 7.36 & 3.89 & 1.96 & 0.359    \\
\textbf{One-D-Piece-B-256} & 4.36 & 2.39 & 1.48 & 0.365   \\
\textbf{One-D-Piece-L-256} & 3.23 & 2.10 & 1.42 & 0.389    \\ \hline
\end{tabular}
}
\caption{\textbf{Comparison of rFID and Linear Probing Accuracy (LPA)} between TiTok models and One-D-Piece models.}
\label{tab:titok_onedpiece_comparison}
\end{table}

\subsection{Downstream Tasks}

\begin{table*}[ht]
\newcommand{\blue}[1]{
  \tikz[baseline]{\node[fill=blue!10,anchor=base] {#1};}
}
\setlength{\dashlinedash}{0.5pt}
\setlength{\dashlinegap}{1.5pt}
\setlength{\arrayrulewidth}{0.5pt}

\centering
\small
\resizebox{\textwidth}{!}{
\begin{tabular}{c|c|ccccc|cc|c}
\hline
\multirow{2}{*}{\textbf{Task}}                  & \multirow{2}{*}{\textbf{Metrics}} & \multicolumn{5}{c|}{\textbf{One-D-Piece-L-256}}   & \multicolumn{2}{c|}{\textbf{Image Formats}} & \multirow{2}{*}{\textbf{Base}}      \\ 
\cline{3-9}
                                       &                        & \textbf{@16} & \textbf{@32} & \textbf{@64} & \textbf{@128} & \textbf{@256}  & \textbf{JPEG} & \textbf{WebP}  &       \\ \hline
\multirow{3}{*}{Object Detection}      & mAP@0.5:0.95$\uparrow$ & 0.051        & 0.097        & \blue{0.180} & \blue{0.264} & \blue{0.305} & 0.001       & 0.166          & ---    \\
                                       & mAP@0.5$\uparrow$      & 0.093        & 0.163        & \blue{0.277} & \blue{0.377} & \blue{0.422} & 0.001       & 0.217          & ---    \\
                                       & mAP@0.75$\uparrow$     & 0.049        & 0.097        & \blue{0.185} & \blue{0.280} & \blue{0.323} & 0.001       & 0.178          & ---    \\ \hline
\multirow{2}{*}{Depth Estimation}      & L1 Loss$\downarrow$    & 2.436        & 1.949        & \blue{1.480} & \blue{1.120} & \blue{0.965} & 3.742       & 1.553          & ---    \\
                                       & L2 Loss$\downarrow$    & 12.282       & 8.593        & 5.326        & \blue{3.160} & \blue{2.362} & 24.097      & 5.050          & ---    \\ \hline
CLIP Emb Reconstruction                & Cos Sim$\uparrow$      & 0.820        & \blue{0.866} & \blue{0.904} & \blue{0.930} & \blue{0.942} & 0.610       & 0.826          & ---    \\ \hline
\multirow{2}{*}{Image Classification}  & Acc@1$\uparrow$        & 0.504        & 0.623        & \blue{0.731} & \blue{0.779} & \blue{0.792} & 0.284       & 0.664          & \textcolor{gray}{0.841}  \\
                                       & Acc@5$\uparrow$        & 0.718        & 0.831        & \blue{0.908} & \blue{0.937} & \blue{0.946} & 0.479       & 0.870          & \textcolor{gray}{0.969}  \\ \hline
\multirow{2}{*}{Semantic Segmentation} & mIoU$\uparrow$         & 0.321        & \blue{0.424} & \blue{0.525} & \blue{0.572} & \blue{0.585} & 0.059       & 0.410          & \textcolor{gray}{0.606}  \\ 
                                       & bIoU $\uparrow$        & 0.146        & 0.210        & \blue{0.281} & \blue{0.315} & \blue{0.325} & 0.027       & 0.211          & \textcolor{gray}{0.343}  \\ \hline
\end{tabular}
}
\caption{\textbf{Evaluation across multiple downstream tasks at different token lengths compared to compression formats}. The scores in the ``Base`` for Image Classification and Semantic Segmentation represent comparisons between the dataset's ground truth and the model's predictions. In contrast, for Object Detection, Depth Estimation, and CLIP Emb Reconstruction, the model's predictions are used as ground truth, as these tasks do not include them in the dataset. \colorbox{blue!10}{Blue background} indicates scores in which One-D-Piece-L-256 surpasses WebP.}
\label{tab:downstream_tasks}
\end{table*}

We evaluate the reconstructed images on various computer vision tasks, including image classification, object detection, semantic segmentation, depth estimation, and CLIP embedding reconstruction, to demonstrate the effectiveness of the tokenizer in real-world applications. This evaluation allows us to assess how well the reconstructed images retain critical information, highlighting the balances between compression efficiency and task performance. \autoref{tab:downstream_tasks} shows the downstream task performance of One-D-Piece-L-256, JPEG and WebP.

\subsubsection{Task Settings}

\noindent \textbf{Image Classification.}
We use the ImageNet validation split and measure Acc@1 and Acc@5 by comparing classification results on reconstructed images. ConvNeXT~\cite{9879745} outputs are used as the ground truth.

\noindent \textbf{Object Detection.}
We use the COCO val2017~\cite{10.1007/978-3-319-10602-1_48} and employ YOLO11x~\cite{UltralyticsYOLO11} as the detection model. Performance is evaluated with mAP@0.5:0.95, mAP@0.5, and mAP@0.75.

\noindent \textbf{Semantic Segmentation.}
We evaluate mean Intersection over Union (mIoU) and boundary IoU (bIoU) on the ImageNet-S~\cite{gao2021luss} dataset, which provides high-quality semantic segmentation annotations based on the ImageNet-1K. SERE~\cite{gao2021luss} is used for the segmentation model.

\noindent \textbf{Depth Estimation.}
We use the ImageNet validation split with Depth Anything~\cite{depthanything} serving as the ground truth. We evaluate the L1 and L2 error of depth estimations between the reconstructed images and the ground truth. 

\noindent \textbf{CLIP Embedding Reconstruction.}
We further assess the quality of semantic reconstruction using CLIP embeddings, with the ImageNet validation split and CLIP~\cite{radford2021learningtransferablevisualmodels} as the ground truth. Cosine similarity is computed between the CLIP embeddings of the reconstructed images and the original images. 

\subsubsection{Results}
As shown in \autoref{tab:downstream_tasks}, One-D-Piece outperforms JPEG across all tasks with only 16 tokens. For tasks where semantic information is crucial, such as CLIP Embedding Reconstruction and Semantic Segmentation, One-D-Piece with 32 or 64 tokens surpasses WebP, achieving a CLIP score of 0.866 versus 0.826 for WebP and an mIoU of 0.424 versus 0.410. In tasks focused on object representation, such as Object Detection and Image Classification, One-D-Piece outperforms WebP with 64 tokens, reaching an mAP@0.5:0.95 of 0.180 versus 0.166 for WebP, and Acc@1 and Acc@5 scores of 0.731 and 0.908, compared to 0.664 and 0.870 for WebP. For Depth Estimation, which requires pixel-level detail, One-D-Piece achieves better L1 and L2 Loss scores at 128 tokens, with values of 1.120 and 3.160, compared to 1.553 and 5.050 for WebP. 

Our results show that One-D-Piece uses only 128 tokens, approximately 10\% of WebP's byte size per image, yet outperforms WebP across all tasks. Its high compression and preserved quality make it ideal for applications like visual question answering and image or video generation.

\subsection{Analysis}
One-D-Piece demonstrates strong performance in reconstruction quality and adaptability for downstream tasks. We attribute this to the introduction of Tail Token Drop regularization. To validate this assumption, we conduct a detailed analysis to uncover the behavior of our models.

\begin{figure*}[htbp]
    \centering
    \begin{minipage}{0.64\linewidth}
        \centering
        \includegraphics[width=\linewidth]{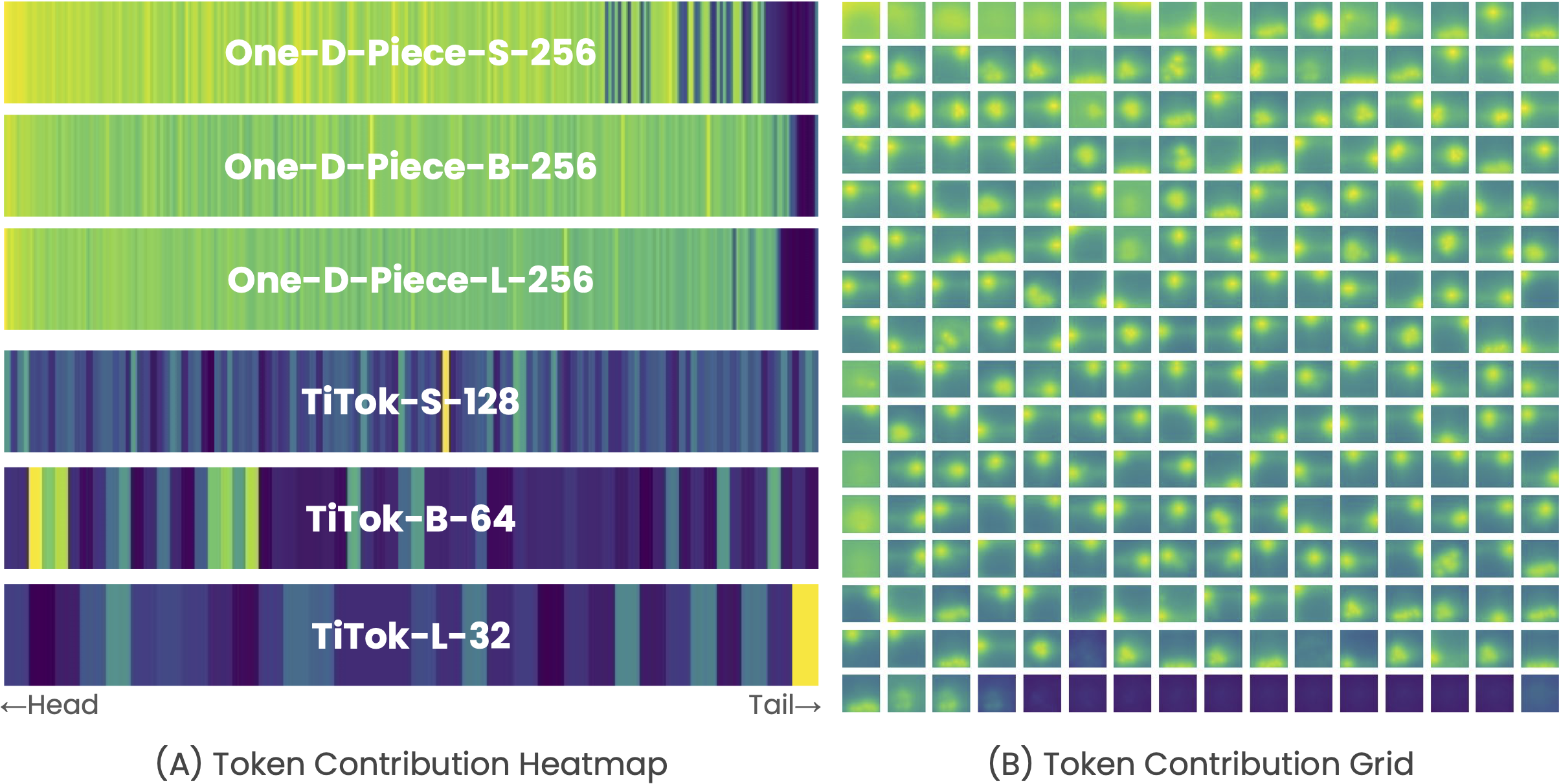}
        \caption{\textbf{The Visualization of Our Token Contribution Analysis.} 
        (A) The head tokens capture global information, as indicated by the strong yellow color, while the later tokens show more localized and weaker peaks.
        (B) Heatmaps of token contributions from the One-D-Piece-L-256 model, displayed in a grid layout for all 256 tokens. Each map highlights the spatial regions to which each token most strongly corresponds. The early tokens capture global features of the entire image, while the mid-to-late tokens respond more strongly to localized, specific regions.
        }
        \label{fig:token_contribution_map}
    \end{minipage}
    \hfill
    \begin{minipage}{0.34\linewidth}
        \centering
        \includegraphics[width=\linewidth]{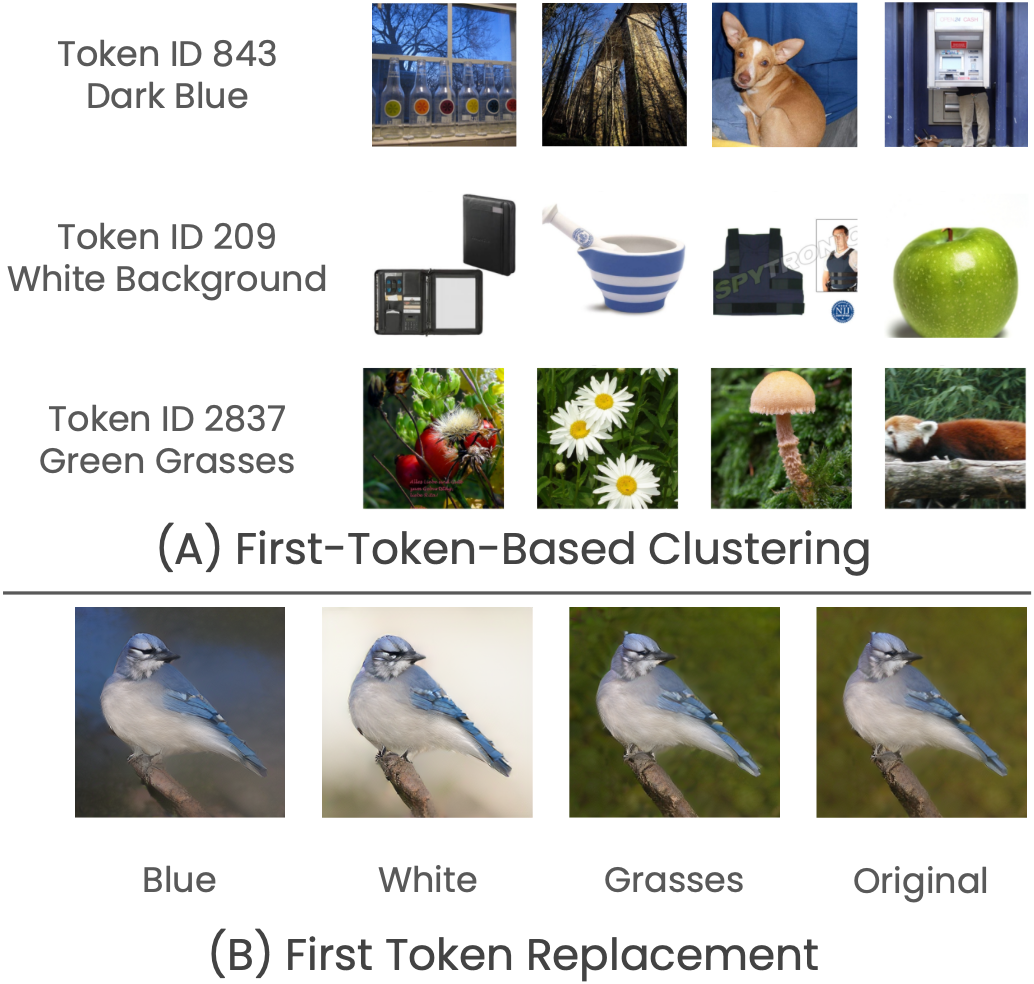}
        \caption{\textbf{The Result of First-Token-based Analysis.} 
        (A) The first tokens correspond to global, especially the background information of the image.
        (B) By replacing the first token to another tokens, we can observe the background changes accordingly, especially when fewer token reconstruction.
        }
        \label{fig:first_token_clustering}
    \end{minipage}
\end{figure*}

\noindent \textbf{Head Tokens Have More Contribution.}
Our Tail Token Drop technique aims to encourage important information to be concentrated at the head of the image token sequence. To verify this hypothesis, we analyze the contribution of each token in the tokenized sequence $\mathbf{q} = [q_1, q_2, \ldots, q_n]$ towards the reconstructed image $\hat{\mathbf{X}}$. Specifically, we perform random replacement for each token $q_i$ in the sequence. Let $\mathbf{q}' = [q_1, \ldots, q_{i-1}, q_i', q_{i+1}, \ldots, q_n]$ be the modified sequence where $q_i$ is replaced by a randomly sampled token $q_i'$. We then generate the reconstructed image $\hat{\mathbf{X}}'$ from the modified sequence $\mathbf{q}'$. To measure the contribution of each token $q_i$, we compute the L1 error between the original reconstruction $\hat{\mathbf{X}}$ and the modified reconstruction $\hat{\mathbf{X}}'$ as $\|\hat{\mathbf{X}}'-\hat{\mathbf{X}}\|_1$. A larger L1 error indicates a greater contribution of the token $q_i$ towards the reconstruction. Furthermore, by retaining the pixel-wise L1 error for each token replacement, we can visualize the spatial regions of the image that are influenced by specific token positions $i$. This allows us to better understand the relationship between token positions in $\mathbf{q}$ and the corresponding pixel areas in $\hat{\mathbf{X}}$. We use ImageNet validation split to examine average behaviors.

The results are shown in \autoref{fig:token_contribution_map} (A). As expected, One-D-Piece tokenizers exhibit a strong peak (yellow) at the head of the token sequence, while TiTok models show mostly random peaks. This confirms that our Tail Token Drop approach effectively encourages the model to concentrate important information at the start of the token sequence, resulting in large contributions.

Interestingly, we observe that some middle and later tokens still correspond to specific spatial regions of the image, despite the tokenizer being trained as a purely 1D tokenizer, as shown in \autoref{fig:token_contribution_map} (B). This suggests that tokens at certain indices retain a strong connection to the 2D structure of the input image, even under the Tail Token Drop constraint.

Additionally, the tokens at the very end of the sequence show almost no contribution, suggesting they carry little meaningful information and that the latent space is not fully utilized. This indicates the potential for creating a more efficient tokenizer by packing more information into currently less important tokens. We leave this exploration as an area for future research.

\noindent \textbf{First Token Encodes Global Information.}
While our analysis of token contribution confirm that more important information is concentrated in the head tokens, the actual content of these information remains unclear. To qualitatively investigate this, we perform clustering by the first token and present the result.

\autoref{fig:first_token_clustering} (A) illustrates ImageNet validation split, clustered by the first token. It can be observed that the first tokens capture the overall similarity between images, indicating that they include global information about the images. Furthermore, we find that replacing the first token leads to corresponding changes in the reconstructed image as shown in \autoref{fig:first_token_clustering} (B) . Although this effect diminishes with longer token sequences, significant influence can be seen in the reconstructions with shorter sequences.

\paragraph{One-D-Piece Tokens are More Semantic.}
Our analysis reveal that Tail Token Drop effectively aggregates global information towards the head of the token sequence, with the leading tokens capturing the overall similarity of the images. To further examine the semantic behavior of One-D-Piece tokens, we conduct an analysis through linear probing experiment, following the MAE protocol~\cite{he2022masked}, which is used in the TiTok report~\cite{yu2024an_titok}. 
Specifically, we append a linear classifier for the output of One-D-Piece encoder and train it for the ImageNet-1K classification task. This evaluation measures the linear separability of the encoded representations, indicating how well the latent features capture semantic information.

As shown in \autoref{tab:titok_onedpiece_comparison}, our model achieves superior linear probing accuracy compared to pretrained TiTok models. Interestingly, this result contrasts with the claim made by TiTok, where models with fewer maximum latent tokens reportedly achieve better accuracy due to stronger constraints in the latent space. We attribute this contrary outcome to our Tail Token Drop technique, which effectively emulates the benefits of a compact latent space by discarding less informative tail tokens. This result further highlights the effectiveness of the Tail Token Drop approach promoting global information aggregation.

\section{Conclusion}
We introduced One-D-Piece, a discrete image tokenizer capable of quality-controllable tokenization, designed to address the trade-off between compression rate and reconstruction quality of existing image tokenization methods that support only fixed number of tokens. Our approach supports dynamic token counts ranging from 1 to 256 by employing the ``Tail Token Drop'' technique, which concentrates critical information at the start of the sequence. This enables high reconstruction quality even with fewer tokens, achieving significant improvements in perceptual quality over conventional methods like JPEG at comparable byte sizes.

Experimental results on ImageNet-1K validate that One-D-Piece not only achieves an rFID of 1.08 with 256 tokens, surpassing prior methods, but also excels in various downstream tasks, including image classification, object detection, and semantic segmentation. Our detailed analysis verified the effectiveness of the Tail Token Drop regularization for encouraging the model to concentrate important information to the head of the sequence. These results highlight the potential of One-D-Piece as an efficient and adaptive tokenizer suitable for a wide range of computer vision applications. By enabling variable-length tokenization, our model sets a new benchmark in compression efficiency and quality, offering promising applications in vision-language models and image and video generation tasks.

\noindent \textbf{Limitations.}
Although the architecture can employ a higher number of tokens, our evaluation was limited to a maximum of 256 tokens. The impact of increasing token count on reconstruction quality and pixel-level performance remains unexplored. Future work will focus on extending the maximum token count to further improve accuracy and quality, as well as developing more advanced training protocols to enhance the preservation of important details.

{
    \small
    \bibliographystyle{ieeenat_fullname}
    \bibliography{main}

\begin{thebibliography}{42}
\providecommand{\natexlab}[1]{#1}
\providecommand{\url}[1]{\texttt{#1}}
\expandafter\ifx\csname urlstyle\endcsname\relax
  \providecommand{\doi}[1]{doi: #1}\else
  \providecommand{\doi}{doi: \begingroup \urlstyle{rm}\Url}\fi

\bibitem[iso(2019)]{iso_jpeg2000_2019}
Information technology — jpeg 2000 image coding system — part 1: Core coding system, 2019.

\bibitem[Bai et~al.(2023)Bai, Bai, Yang, Wang, Tan, Wang, Lin, Zhou, and Zhou]{Qwen-VL}
Jinze Bai, Shuai Bai, Shusheng Yang, Shijie Wang, Sinan Tan, Peng Wang, Junyang Lin, Chang Zhou, and Jingren Zhou.
\newblock Qwen-vl: A versatile vision-language model for understanding, localization, text reading, and beyond.
\newblock \emph{arXiv preprint arXiv:2308.12966}, 2023.

\bibitem[Blau and Michaeli(2019)]{blau2019rethinking}
Yochai Blau and Tomer Michaeli.
\newblock Rethinking lossy compression: The rate-distortion-perception tradeoff.
\newblock In \emph{International Conference on Machine Learning}, pages 675--685. PMLR, 2019.

\bibitem[Bruce et~al.(2024)Bruce, Dennis, Edwards, Parker-Holder, Shi, Hughes, Lai, Mavalankar, Steigerwald, Apps, et~al.]{bruce2024genie}
Jake Bruce, Michael~D Dennis, Ashley Edwards, Jack Parker-Holder, Yuge Shi, Edward Hughes, Matthew Lai, Aditi Mavalankar, Richie Steigerwald, Chris Apps, et~al.
\newblock Genie: Generative interactive environments.
\newblock In \emph{Forty-first International Conference on Machine Learning}, 2024.

\bibitem[Chang et~al.(2022)Chang, Zhang, Jiang, Liu, and Freeman]{chang2022maskgit}
Huiwen Chang, Han Zhang, Lu Jiang, Ce Liu, and William~T Freeman.
\newblock Maskgit: Masked generative image transformer.
\newblock In \emph{Proceedings of the IEEE/CVF Conference on Computer Vision and Pattern Recognition}, pages 11315--11325, 2022.

\bibitem[Deng et~al.(2009)Deng, Dong, Socher, Li, Li, and Fei-Fei]{deng2009imagenet}
Jia Deng, Wei Dong, Richard Socher, Li-Jia Li, Kai Li, and Li Fei-Fei.
\newblock Imagenet: A large-scale hierarchical image database.
\newblock In \emph{2009 IEEE Conference on Computer Vision and Pattern Recognition}, pages 248--255, 2009.

\bibitem[Dhariwal and Nichol(2024)]{10.5555/3540261.3540933}
Prafulla Dhariwal and Alex Nichol.
\newblock Diffusion models beat gans on image synthesis.
\newblock In \emph{Proceedings of the 35th International Conference on Neural Information Processing Systems}, Red Hook, NY, USA, 2024. Curran Associates Inc.

\bibitem[Dosovitskiy et~al.(2021)Dosovitskiy, Beyer, Kolesnikov, Weissenborn, Zhai, Unterthiner, Dehghani, Minderer, Heigold, Gelly, Uszkoreit, and Houlsby]{dosovitskiy2021vit}
Alexey Dosovitskiy, Lucas Beyer, Alexander Kolesnikov, Dirk Weissenborn, Xiaohua Zhai, Thomas Unterthiner, Mostafa Dehghani, Matthias Minderer, Georg Heigold, Sylvain Gelly, Jakob Uszkoreit, and Neil Houlsby.
\newblock An image is worth 16x16 words: Transformers for image recognition at scale.
\newblock In \emph{International Conference on Learning Representations}, 2021.

\bibitem[Esser et~al.(2021)Esser, Rombach, and Ommer]{esser2021taming}
Patrick Esser, Robin Rombach, and Bj{\"o}rn Ommer.
\newblock Taming transformers for high-resolution image synthesis.
\newblock In \emph{2021 IEEE/CVF Conference on Computer Vision and Pattern Recognition (CVPR)}, pages 12868--12878. IEEE, 2021.

\bibitem[Gao et~al.(2021)Gao, Li, Yang, Cheng, Han, and Torr]{gao2021luss}
Shanghua Gao, Zhong-Yu Li, Ming-Hsuan Yang, Ming-Ming Cheng, Junwei Han, and Philip Torr.
\newblock Large-scale unsupervised semantic segmentation.
\newblock \emph{arXiv preprint arXiv:2106.03149}, 2021.

\bibitem[Ge et~al.(2024{\natexlab{a}})Ge, Zhao, Zeng, Ge, Li, Wang, and Shan]{ge2024making_seed_tokenizer}
Yuying Ge, Sijie Zhao, Ziyun Zeng, Yixiao Ge, Chen Li, Xintao Wang, and Ying Shan.
\newblock Making {LL}a{MA} {SEE} and draw with {SEED} tokenizer.
\newblock In \emph{The Twelfth International Conference on Learning Representations}, 2024{\natexlab{a}}.

\bibitem[Ge et~al.(2024{\natexlab{b}})Ge, Zhao, Zhu, Ge, Yi, Song, Li, Ding, and Shan]{ge2024seed}
Yuying Ge, Sijie Zhao, Jinguo Zhu, Yixiao Ge, Kun Yi, Lin Song, Chen Li, Xiaohan Ding, and Ying Shan.
\newblock Seed-x: Multimodal models with unified multi-granularity comprehension and generation.
\newblock \emph{arXiv preprint arXiv:2404.14396}, 2024{\natexlab{b}}.

\bibitem[Google(2024)]{webp}
Google.
\newblock {WebP}: Compression techniques, 2024.
\newblock Accessed: 2024-11-11.

\bibitem[He et~al.(2022)He, Chen, Xie, Li, Doll{\'a}r, and Girshick]{he2022masked}
Kaiming He, Xinlei Chen, Saining Xie, Yanghao Li, Piotr Doll{\'a}r, and Ross Girshick.
\newblock Masked autoencoders are scalable vision learners.
\newblock In \emph{Proceedings of the IEEE/CVF conference on computer vision and pattern recognition}, pages 16000--16009, 2022.

\bibitem[Heusel et~al.(2017)Heusel, Ramsauer, Unterthiner, Nessler, and Hochreiter]{heusel2017gans}
Martin Heusel, Hubert Ramsauer, Thomas Unterthiner, Bernhard Nessler, and Sepp Hochreiter.
\newblock Gans trained by a two time-scale update rule converge to a local nash equilibrium.
\newblock \emph{Advances in neural information processing systems}, 30, 2017.

\bibitem[Jocher and Qiu(2024)]{UltralyticsYOLO11}
Glenn Jocher and Jing Qiu.
\newblock {Ultralytics YOLO11}.
\newblock \url{https://docs.ultralytics.com/models/yolo11/}, 2024.

\bibitem[Kingma and Welling(2014)]{Kingma2014_vae}
Diederik~P. Kingma and Max Welling.
\newblock {Auto-Encoding Variational Bayes}.
\newblock In \emph{2nd International Conference on Learning Representations, {ICLR} 2014, Banff, AB, Canada, April 14-16, 2014, Conference Track Proceedings}, 2014.

\bibitem[Koike-Akino and Wang(2020)]{koike_wang_2020_taildrop}
Toshiaki Koike-Akino and Ye Wang.
\newblock Stochastic bottleneck: Rateless auto-encoder for flexible dimensionality reduction.
\newblock In \emph{2020 IEEE International Symposium on Information Theory (ISIT)}, pages 2735--2740, 2020.

\bibitem[Kudo and Richardson(2018)]{kudo2018sentencepiecesimplelanguageindependent}
Taku Kudo and John Richardson.
\newblock Sentencepiece: A simple and language independent subword tokenizer and detokenizer for neural text processing, 2018.

\bibitem[Lauren{\c{c}}on et~al.(2024)Lauren{\c{c}}on, Tronchon, Cord, and Sanh]{laurencon2024what}
Hugo Lauren{\c{c}}on, Leo Tronchon, Matthieu Cord, and Victor Sanh.
\newblock What matters when building vision-language models?
\newblock In \emph{The Thirty-eighth Annual Conference on Neural Information Processing Systems}, 2024.

\bibitem[Lin et~al.(2014)Lin, Maire, Belongie, Hays, Perona, Ramanan, Doll{\'a}r, and Zitnick]{10.1007/978-3-319-10602-1_48}
Tsung-Yi Lin, Michael Maire, Serge Belongie, James Hays, Pietro Perona, Deva Ramanan, Piotr Doll{\'a}r, and C.~Lawrence Zitnick.
\newblock Microsoft coco: Common objects in context.
\newblock In \emph{Computer Vision -- ECCV 2014}, pages 740--755, Cham, 2014. Springer International Publishing.

\bibitem[Liu et~al.(2024)Liu, Li, Li, Li, Zhang, Shen, and Lee]{liu2024llavanext}
Haotian Liu, Chunyuan Li, Yuheng Li, Bo Li, Yuanhan Zhang, Sheng Shen, and Yong~Jae Lee.
\newblock Llava-next: Improved reasoning, ocr, and world knowledge, 2024.

\bibitem[Liu et~al.(2022)Liu, Mao, Wu, Feichtenhofer, Darrell, and Xie]{9879745}
Zhuang Liu, Hanzi Mao, Chao-Yuan Wu, Christoph Feichtenhofer, Trevor Darrell, and Saining Xie.
\newblock A convnet for the 2020s.
\newblock In \emph{2022 IEEE/CVF Conference on Computer Vision and Pattern Recognition (CVPR)}, pages 11966--11976, 2022.

\bibitem[Luo et~al.(2024)Luo, Shi, Ge, Yang, Wang, and Shan]{luo2024open_openmagvitv2}
Zhuoyan Luo, Fengyuan Shi, Yixiao Ge, Yujiu Yang, Limin Wang, and Ying Shan.
\newblock Open-magvit2: An open-source project toward democratizing auto-regressive visual generation.
\newblock \emph{arXiv preprint arXiv:2409.04410}, 2024.

\bibitem[Radford et~al.(2021)Radford, Kim, Hallacy, Ramesh, Goh, Agarwal, Sastry, Askell, Mishkin, Clark, Krueger, and Sutskever]{radford2021learningtransferablevisualmodels}
Alec Radford, Jong~Wook Kim, Chris Hallacy, Aditya Ramesh, Gabriel Goh, Sandhini Agarwal, Girish Sastry, Amanda Askell, Pamela Mishkin, Jack Clark, Gretchen Krueger, and Ilya Sutskever.
\newblock Learning transferable visual models from natural language supervision, 2021.

\bibitem[Shannon(1959)]{shannon1959coding}
Claude~E. Shannon.
\newblock Coding theorems for a discrete source with a fidelity criterion.
\newblock In \emph{IRE National Convention Record}, pages 142--163, 1959.

\bibitem[Simonyan and Zisserman(2014)]{simonyan2014very}
Karen Simonyan and Andrew Zisserman.
\newblock Very deep convolutional networks for large-scale image recognition.
\newblock \emph{arXiv preprint arXiv:1409.1556}, 2014.

\bibitem[Sun et~al.(2024)Sun, Jiang, Chen, Zhang, Peng, Luo, and Yuan]{sun2024autoregressive}
Peize Sun, Yi Jiang, Shoufa Chen, Shilong Zhang, Bingyue Peng, Ping Luo, and Zehuan Yuan.
\newblock Autoregressive model beats diffusion: Llama for scalable image generation.
\newblock \emph{arXiv preprint arXiv:2406.06525}, 2024.

\bibitem[Team(2024)]{chameleonteam2024chameleonmixedmodalearlyfusionfoundation}
Chameleon Team.
\newblock Chameleon: Mixed-modal early-fusion foundation models, 2024.

\bibitem[van~den Oord et~al.(2017)van~den Oord, Vinyals, and kavukcuoglu]{NIPS2017_7a98af17_vqvae}
Aaron van~den Oord, Oriol Vinyals, and koray kavukcuoglu.
\newblock Neural discrete representation learning.
\newblock In \emph{Advances in Neural Information Processing Systems}. Curran Associates, Inc., 2017.

\bibitem[Vaswani et~al.(2017)Vaswani, Shazeer, Parmar, Uszkoreit, Jones, Gomez, Kaiser, and Polosukhin]{vaswani2017_attention}
Ashish Vaswani, Noam Shazeer, Niki Parmar, Jakob Uszkoreit, Llion Jones, Aidan~N Gomez, \L~ukasz Kaiser, and Illia Polosukhin.
\newblock Attention is all you need.
\newblock In \emph{Advances in Neural Information Processing Systems}. Curran Associates, Inc., 2017.

\bibitem[Villegas et~al.(2022)Villegas, Babaeizadeh, Kindermans, Moraldo, Zhang, Saffar, Castro, Kunze, and Erhan]{villegas2022phenaki}
Ruben Villegas, Mohammad Babaeizadeh, Pieter-Jan Kindermans, Hernan Moraldo, Han Zhang, Mohammad~Taghi Saffar, Santiago Castro, Julius Kunze, and Dumitru Erhan.
\newblock Phenaki: Variable length video generation from open domain textual descriptions.
\newblock In \emph{International Conference on Learning Representations}, 2022.

\bibitem[Wallace(1991)]{wallace1991jpeg}
Gregory~K Wallace.
\newblock The jpeg still picture compression standard.
\newblock \emph{Communications of the ACM}, 34\penalty0 (4):\penalty0 30--44, 1991.

\bibitem[Wang et~al.(2024{\natexlab{a}})Wang, Jiang, Yuan, Peng, Wu, and Jiang]{wang2024omnitokenizer}
Junke Wang, Yi Jiang, Zehuan Yuan, Binyue Peng, Zuxuan Wu, and Yu-Gang Jiang.
\newblock Omnitokenizer: A joint image-video tokenizer for visual generation.
\newblock \emph{arXiv preprint arXiv:2406.09399}, 2024{\natexlab{a}}.

\bibitem[Wang et~al.(2024{\natexlab{b}})Wang, Zhang, Luo, Sun, Cui, Wang, Zhang, Wang, Li, Yu, et~al.]{wang2024emu3}
Xinlong Wang, Xiaosong Zhang, Zhengxiong Luo, Quan Sun, Yufeng Cui, Jinsheng Wang, Fan Zhang, Yueze Wang, Zhen Li, Qiying Yu, et~al.
\newblock Emu3: Next-token prediction is all you need.
\newblock \emph{arXiv preprint arXiv:2409.18869}, 2024{\natexlab{b}}.

\bibitem[Weber et~al.(2024)Weber, Yu, Yu, Deng, Shen, Cremers, and Chen]{weber2024maskbit}
Mark Weber, Lijun Yu, Qihang Yu, Xueqing Deng, Xiaohui Shen, Daniel Cremers, and Liang-Chieh Chen.
\newblock Maskbit: Embedding-free image generation via bit tokens.
\newblock \emph{arXiv preprint arXiv:2409.16211}, 2024.

\bibitem[Wu et~al.(2016)Wu, Schuster, Chen, Le, Norouzi, Macherey, Krikun, Cao, Gao, Macherey, et~al.]{wu2016google}
Yonghui Wu, Mike Schuster, Zhifeng Chen, Quoc~V Le, Mohammad Norouzi, Wolfgang Macherey, Maxim Krikun, Yuan Cao, Qin Gao, Klaus Macherey, et~al.
\newblock Google's neural machine translation system: Bridging the gap between human and machine translation.
\newblock \emph{arXiv preprint arXiv:1609.08144}, 2016.

\bibitem[Yan et~al.(2021)Yan, Zhang, Abbeel, and Srinivas]{yan2021videogpt}
Wilson Yan, Yunzhi Zhang, Pieter Abbeel, and Aravind Srinivas.
\newblock Videogpt: Video generation using vq-vae and transformers, 2021.

\bibitem[Yang et~al.(2024)Yang, Kang, Huang, Xu, Feng, and Zhao]{depthanything}
Lihe Yang, Bingyi Kang, Zilong Huang, Xiaogang Xu, Jiashi Feng, and Hengshuang Zhao.
\newblock Depth anything: Unleashing the power of large-scale unlabeled data.
\newblock In \emph{CVPR}, 2024.

\bibitem[Yu et~al.(2024{\natexlab{a}})Yu, Lezama, Gundavarapu, Versari, Sohn, Minnen, Cheng, Gupta, Gu, Hauptmann, Gong, Yang, Essa, Ross, and Jiang]{yu2024language_magvitv2}
Lijun Yu, Jose Lezama, Nitesh~Bharadwaj Gundavarapu, Luca Versari, Kihyuk Sohn, David Minnen, Yong Cheng, Agrim Gupta, Xiuye Gu, Alexander~G Hauptmann, Boqing Gong, Ming-Hsuan Yang, Irfan Essa, David~A Ross, and Lu Jiang.
\newblock Language model beats diffusion - tokenizer is key to visual generation.
\newblock In \emph{The Twelfth International Conference on Learning Representations}, 2024{\natexlab{a}}.

\bibitem[Yu et~al.(2024{\natexlab{b}})Yu, Weber, Deng, Shen, Cremers, and Chen]{yu2024an_titok}
Qihang Yu, Mark Weber, Xueqing Deng, Xiaohui Shen, Daniel Cremers, and Liang-Chieh Chen.
\newblock An image is worth 32 tokens for reconstruction and generation.
\newblock In \emph{Advances in Neural Information Processing Systems}, 2024{\natexlab{b}}.

\bibitem[Zhang et~al.(2018)Zhang, Isola, Efros, Shechtman, and Wang]{zhang2018perceptual}
Richard Zhang, Phillip Isola, Alexei~A Efros, Eli Shechtman, and Oliver Wang.
\newblock The unreasonable effectiveness of deep features as a perceptual metric.
\newblock In \emph{CVPR}, 2018.

\end{thebibliography}
}

\clearpage
\appendix
\setcounter{page}{1}
\setcounter{section}{0}

\section{Training Details}

For our experiments, we strictly adhered to the TiTok settings~\cite{yu2024an_titok} as detailed in \autoref{tab:hyperparameters} for both model architecture and training configurations. As the base implementation of TiTok, we utilized the repository \href{https://github.com/bytedance/1d-tokenizer/tree/6cf0d6e63a339eede815cec2d2af6cb622f2e3fe}{bytedance/1d-tokenizer}, referencing commit ID \texttt{6cf0d6}, following its \href{https://github.com/bytedance/1d-tokenizer/blob/6cf0d6e63a339eede815cec2d2af6cb622f2e3fe/LICENSE}{Apache 2.0 License}.

\section{Evaluation Details}

\subsection{Reconstruction}
\begin{figure*}[ht]
    \centering
    \includegraphics[width=1.0\linewidth]{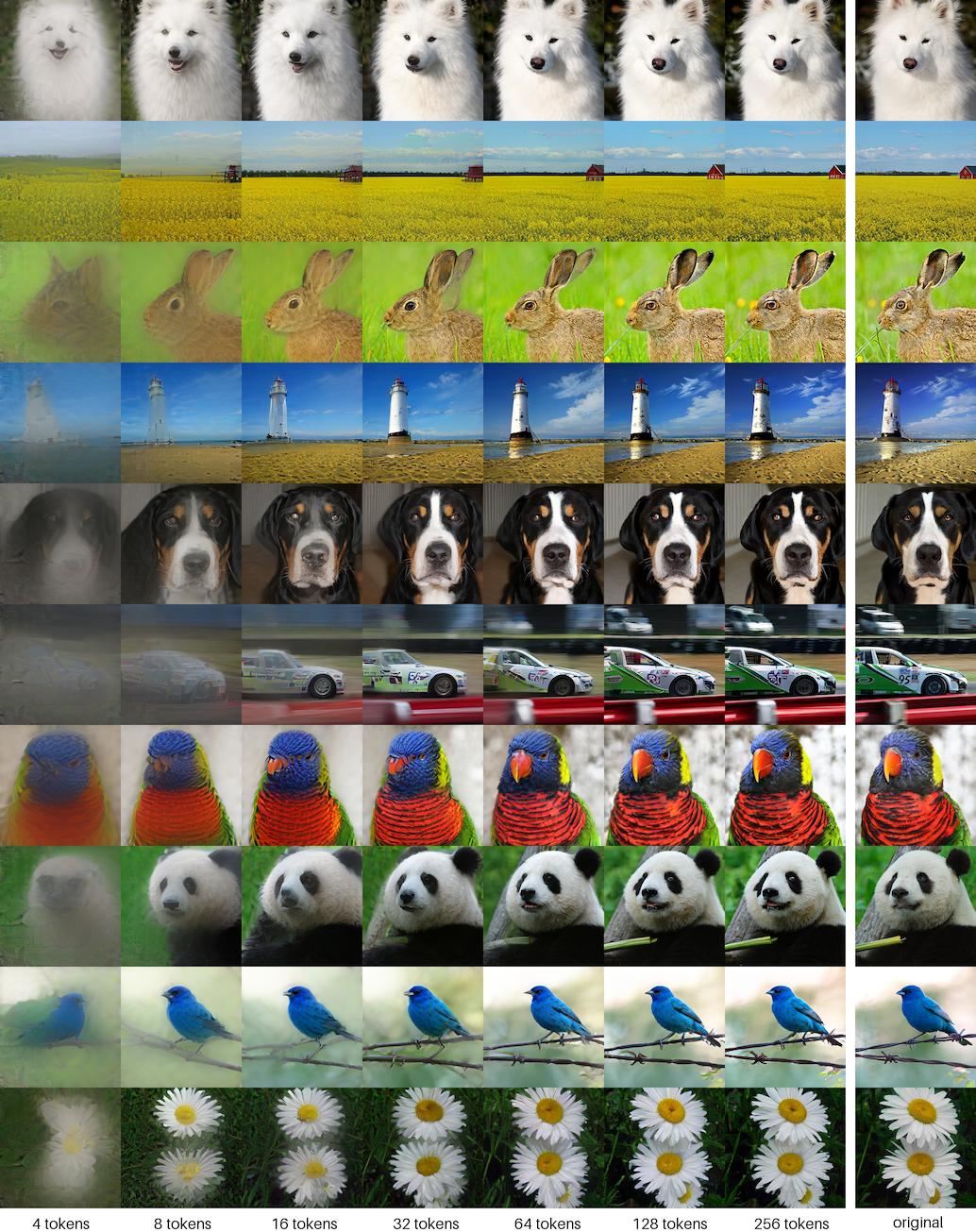}
    \caption{\textbf{Visual comparison of reconstructed images with One-D-Piece-L-256}.}
    \label{fig:L-256-reconstruction_image}
\end{figure*}

\begin{figure*}[ht]
    \centering
    \includegraphics[width=1.0\linewidth]{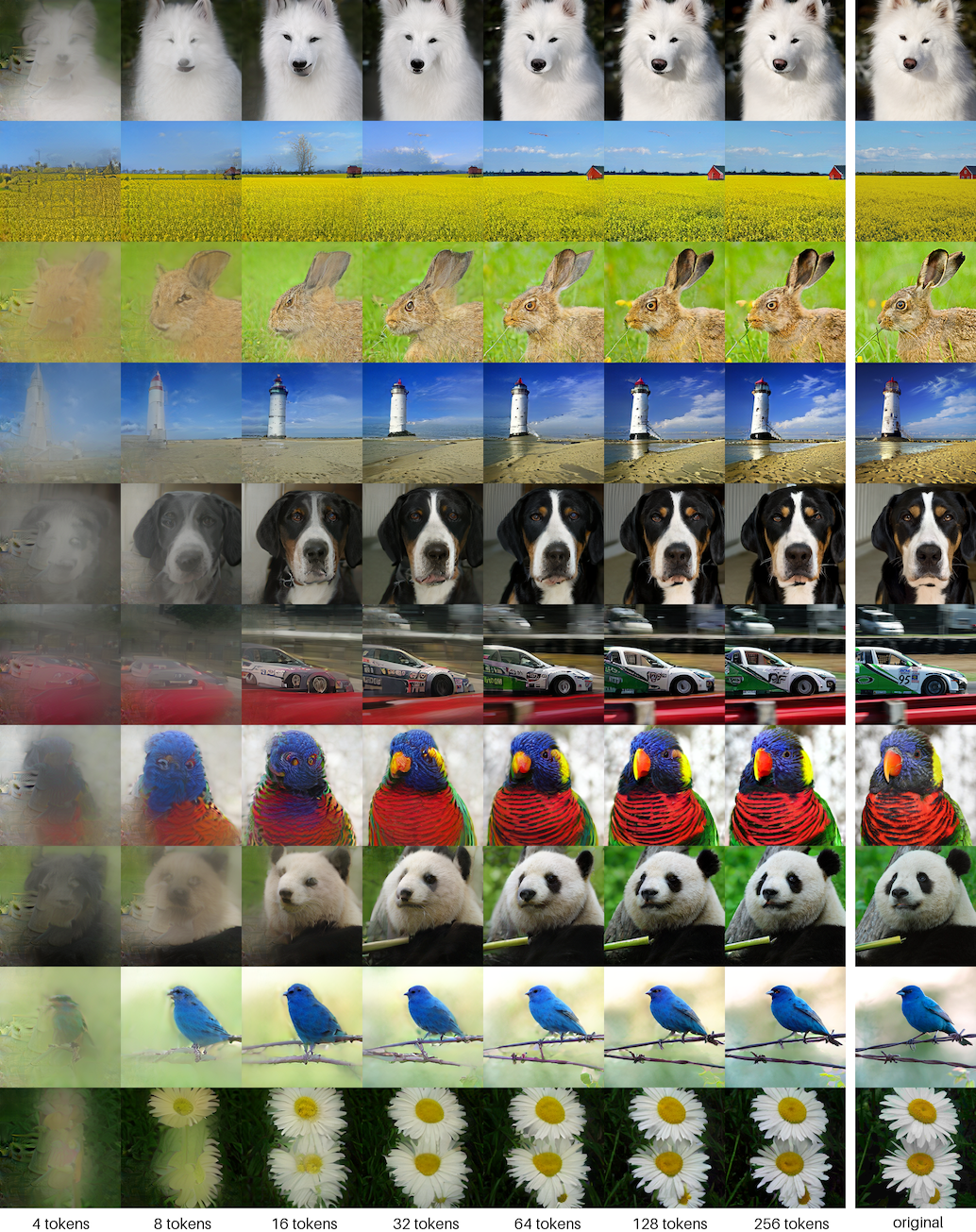}
    \caption{\textbf{Visual comparison of reconstructed images with One-D-Piece-B-256}.}
    \label{fig:B-256-reconstruction_image}
\end{figure*}

\begin{figure*}[ht]
    \centering
    \includegraphics[width=1.0\linewidth]{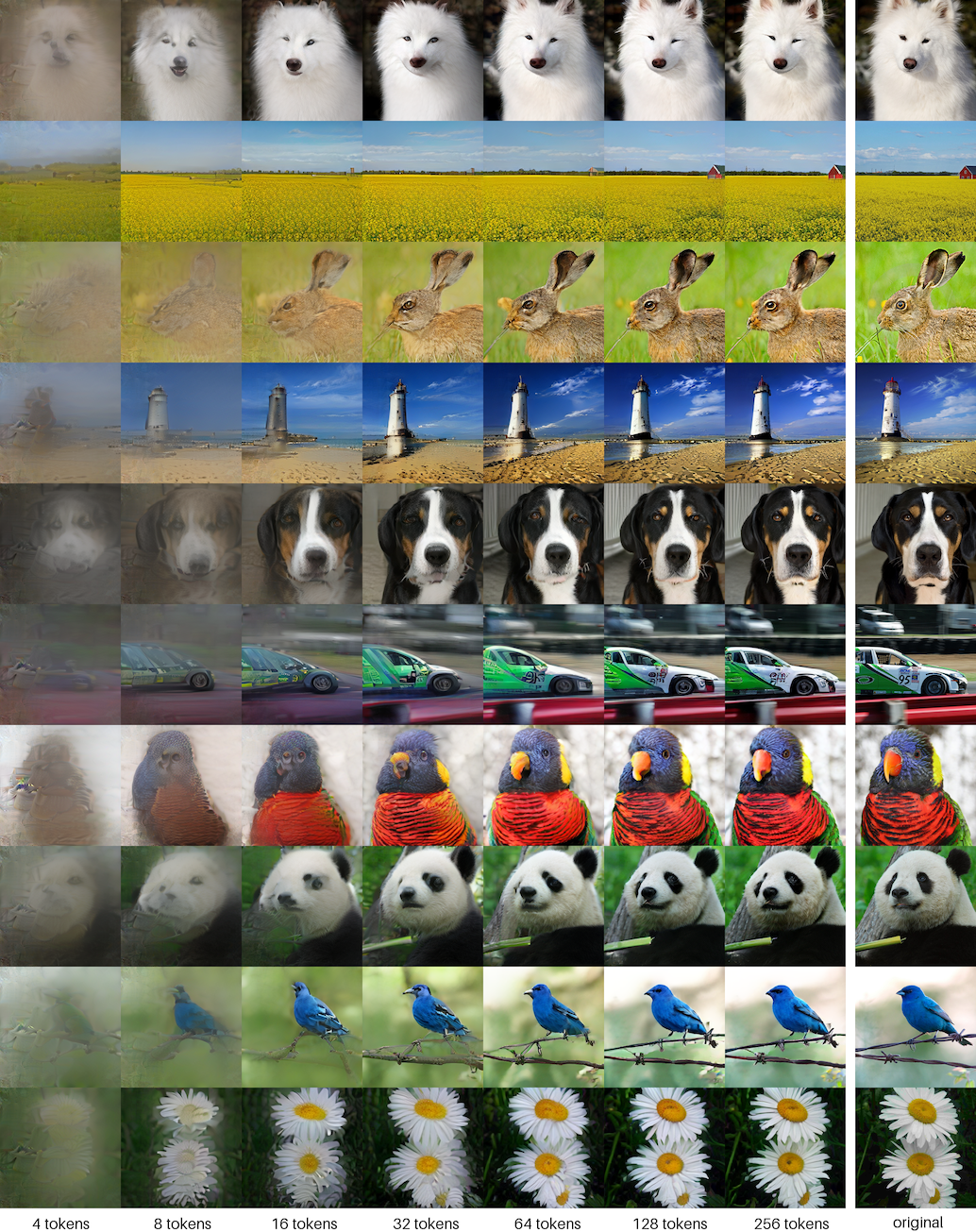}
    \caption{\textbf{Visual comparison of reconstructed images with One-D-Piece-S-256}.}
    \label{fig:S-256-reconstruction_image}
\end{figure*}

\paragraph{Additional Samples.}We further provide sample reconstructed images, where L-256, B-256, and S-256 are shown in \autoref{fig:L-256-reconstruction_image}, \autoref{fig:B-256-reconstruction_image}, and \autoref{fig:S-256-reconstruction_image}, respectively.

\paragraph{rFID Values.} We provide the detailed rFID values corresponding to \autoref{fig:length_to_rfid} in \autoref{tab:compression_quality}. JPEG, JPEG 2000, and WebP format files were generated using OpenCV version 4.10.0.84 as the converter.
\begin{table}[h!]
\centering
\resizebox{\linewidth}{!}{
\begin{tabular}{lcccc}
\toprule
\textbf{Method} & \textbf{Quality} & \textbf{BpI} & \textbf{Tokens} & \textbf{rFID}$\downarrow$ \\
\midrule
\multirow{6}{*}{JPEG} 
    & 1  & 2063.30  & 1376 & 113.30 \\
    & 2  & 2064.39  & 1376 & 113.24 \\
    & 4  & 2374.21  & 1583 & 79.23 \\
    & 8  & 3243.59  & 2167 & 32.92 \\
    & 16 & 4749.32  & 3167 & 15.76 \\
    & 24 & 5995.47  & 3997 & 12.12 \\
\midrule
\multirow{6}{*}{JPEG 2000}
    & 2  & 406.30   & 271  & 299.40 \\
    & 4  & 795.08   & 531  & 139.99  \\
    & 8  & 1573.19  & 1049 & 79.90  \\
    & 16 & 3130.82  & 2088 & 43.23 \\
    & 24 & 4694.12  & 3130 & 26.40  \\
    & 30 & 5869.50  & 3913 & 19.09 \\
\midrule
\multirow{6}{*}{WebP}
    & 0  & 1964.02 & 1310 & 31.98 \\
    & 2  & 2206.52 & 1472 & 28.81 \\
    & 4  & 2575.20 & 1717 & 25.05 \\
    & 8  & 3136.88 & 2729 & 16.26  \\
    & 16 & 4029.67 & 2729 & 16.26  \\
    & 32 & 5825.86 & 3884 & 11.47 \\
\bottomrule
\end{tabular}
}
\caption{Comparison of image quality against compression quality for JPEG, JPEG 2000, and WebP. BpI stands for Byte-per-Image.}
\label{tab:compression_quality}
\end{table}

\subsection{Downstream Tasks}
In addition to the reported result of One-D-Piece-L-256, we further provide results for other variants. The evaluation results for downstream tasks using One-D-Piece-S-256 and B-256 are presented in \autoref{tab:downstream_tasks_sb256}. Consistent with our report in the main paper, tasks that rely primarily on semantic information, such as semantic segmentation and CLIP embedding reconstruction, achieve scores that surpass WebP with a smaller number of tokens. Similarly, object detection and image classification tend to show better results with a moderate number of tokens. For the depth estimation task, both models require 128 tokens to outperform WebP.

\subsection{Generation}

Our primary evaluations of One-D-Piece focus on reconstruction quality, but generation quality is also an essential metric, particularly for applications like image and video generation models. Since TiTok~\cite{yu2024an_titok}, the base architecture of One-D-Piece, is reported to contribute to high generation quality, our objective is to confirm that the introduction of the Tail Token Drop mechanism does not adversely affect generation quality.

\begin{figure*}[htbp]
    \centering
    \includegraphics[width=0.9\linewidth]{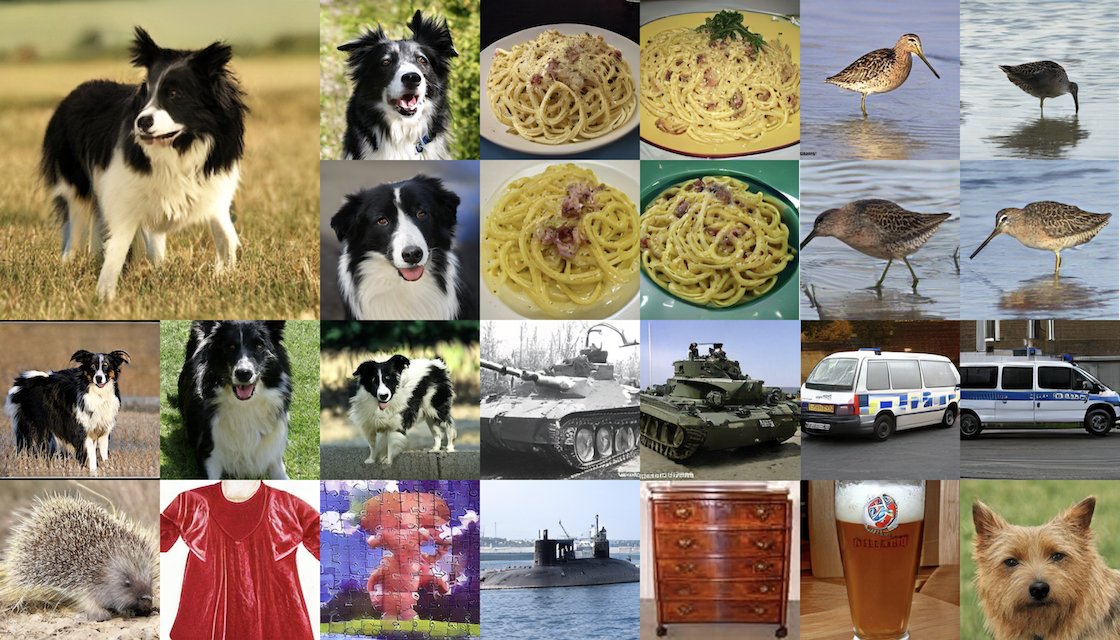}
    \caption{Images generated by MaskGIT with One-D-Piece-L-256 with random classes.}
    \label{fig:generated_results_l256}
\end{figure*}

For the evaluation, we train MaskGIT~\cite{esser2021taming} for class-conditioned image generation, utilizing One-D-Piece as the image tokenizer, following the protocols utilized in TiTok. We assess the generation FID (gFID) using precomputed statistics from the Ablated Diffusion Model~\cite{10.5555/3540261.3540933}. As shown in \autoref{tab:generation_quality}, our models demonstrate competitive performance compared to TiTok variants. This result confirms that the introduction of Tail Token Drop does not damage the generation quality and highlight the potential of One-D-Piece for image and video generation tasks. Our detailed setting for the generation model is shown in \autoref{tab:generation_hyperparameters}. Sample generated images are shown in \autoref{fig:generated_results_l256}.

\begin{table}[h!]
\resizebox{\linewidth}{!}{
\centering
\begin{tabular}{ccccccc}
\hline
\multirow{2}{*}{\textbf{Model}}& \multicolumn{3}{c}{\textbf{TiTok}} & \multicolumn{3}{c}{\textbf{One-D-Piece}} \\ \cmidrule(lr){2-4} \cmidrule(lr){5-7}
                          &\textbf{S-128}&\textbf{B-64}&\textbf{L-32}& \textbf{S-256}  &\textbf{B-256}&\textbf{L-256} \\ \hline 
\textbf{gFID}$\downarrow$ &   2.50       &   2.48      & 2.77        & 2.67            &  2.70        & \textbf{2.35} \\ 
\textbf{IS}$\uparrow$     & ---          & 216.61      & 201.85      & \textbf{265.82} & 259.27       & 224.38 \\ \hline
\end{tabular}
}
\caption{Generation quality comparison between TiTok and One-D-Piece models, evaluated using gFID (lower is better) and Inception Score (IS, higher is better). The Inception Score for TiTok-S-128 is not reported, as the pretrained MaskGIT with Vision Transformer backbone model for this configuration has not been released.}
\label{tab:generation_quality}
\end{table}

\begin{table*}[t]
\newcommand{\green}[1]{
  \tikz[baseline]{\node[fill=green!10,anchor=base] {#1};}
}
\newcommand{\yellow}[1]{
  \tikz[baseline]{\node[fill=yellow!20,anchor=base] {#1};}
}
\setlength{\dashlinedash}{0.5pt}
\setlength{\dashlinegap}{1.5pt} 
\setlength{\arrayrulewidth}{0.5pt}

\centering
\small
\resizebox{0.93\textwidth}{!}{
\begin{tabular}{c|c|ccccc|cc|c}
\hline
\multirow{2}{*}{\textbf{Task}}                  & \multirow{2}{*}{\textbf{Metrics}} & \multicolumn{5}{c|}{\textbf{One-D-Piece-S-256}}   & \multicolumn{2}{c|}{\textbf{Image Formats}} & \multirow{2}{*}{\textbf{Base}}      \\ 
\cline{3-9}
                                       &                        & \textbf{@16} & \textbf{@32} & \textbf{@64} & \textbf{@128} & \textbf{@256} & \textbf{JPEG} & \textbf{WebP}  &       \\ \hline
\multirow{3}{*}{Object Detection}      & mAP@0.5:0.95$\uparrow$ & 0.030        & 0.063        & 0.125        & \green{0.204}  & \green{0.244}  & 0.001         & 0.166          & ---    \\
                                       & mAP@0.5$\uparrow$      & 0.062        & 0.112        & 0.197        & \green{0.300}  & \green{0.349}  & 0.001         & 0.217          & ---    \\
                                       & mAP@0.75$\uparrow$     & 0.025        & 0.061        & 0.129        & \green{0.214}  & \green{0.260}  & 0.001         & 0.178          & ---    \\ \hline
\multirow{2}{*}{Depth Estimation}      & L1 Loss$\downarrow$    & 2.919        & 2.364        & 1.861        & \green{1.482}  & \green{1.340}  & 3.742         & 1.553          & ---    \\
                                       & L2 Loss$\downarrow$    & 16.006       & 11.478       & 7.590        & \green{5.024}  & \green{4.171}  & 24.097        & 5.050          & ---    \\ \hline
CLIP Emb Reconstruction                & Cos Sim$\uparrow$      & 0.779        & 0.832        & \green{0.879} & \green{0.914}  & \green{0.926}  & 0.610         & 0.826          & ---    \\ \hline
\multirow{2}{*}{Image Classification}  & Acc@1$\uparrow$        & 0.378        & 0.535        & 0.659        & \green{0.738}  & \green{0.759}  & 0.284         & 0.664          & \textcolor{gray}{0.841}  \\
                                       & Acc@5$\uparrow$        & 0.613        & 0.769        & 0.864        & \green{0.915}  & \green{0.928}  & 0.479         & 0.870          & \textcolor{gray}{0.969}  \\ \hline
\multirow{2}{*}{Semantic Segmentation} & mIoU$\uparrow$         & 0.2016       & 0.329        & \green{0.438} & \green{0.518}  & \green{0.540}  & 0.059         & 0.410          & \textcolor{gray}{0.606}  \\ 
                                       & bIoU $\uparrow$        & 0.084        & 0.154        & \green{0.223} & \green{0.278}  & \green{0.295}  & 0.027         & 0.211          & \textcolor{gray}{0.343}  \\ \hline
\end{tabular}
}
\vspace{1em}

\resizebox{0.93\textwidth}{!}{
\begin{tabular}{c|c|ccccc|cc|c}
\hline
\multirow{2}{*}{\textbf{Task}}                  & \multirow{2}{*}{\textbf{Metrics}} & \multicolumn{5}{c|}{\textbf{One-D-Piece-B-256}}   & \multicolumn{2}{c|}{\textbf{Image Formats}} & \multirow{2}{*}{\textbf{Base}} \\ 
\cline{3-9}
                                       &                        & \textbf{@16} & \textbf{@32} & \textbf{@64} & \textbf{@128} & \textbf{@256}  & \textbf{JPEG} & \textbf{WebP}  &        \\ \hline
\multirow{3}{*}{Object Detection}      & mAP@0.5:0.95$\uparrow$ & 0.038        & 0.080        & 0.148        & \yellow{0.228}  & \yellow{0.277}   & 0.001         & 0.166          & ---    \\
                                       & mAP@0.5$\uparrow$      & 0.076        & 0.140        & \yellow{0.234} & \yellow{0.337}  & \yellow{0.391}   & 0.001         & 0.217          & ---    \\
                                       & mAP@0.75$\uparrow$     & 0.034        & 0.079        & 0.152        & \yellow{0.235}  & \yellow{0.292}   & 0.001         & 0.178          & ---    \\ \hline
\multirow{2}{*}{Depth Estimation}      & L1 Loss$\downarrow$    & 2.709        & 2.182        & 1.723        & \yellow{1.377}  & \yellow{1.214}   & 3.742         & 1.553          & ---    \\
                                       & L2 Loss$\downarrow$    & 14.404       & 10.095       & 6.682        & \yellow{4.411}  & \yellow{3.491}   & 24.097        & 5.050          & ---    \\ \hline
CLIP Emb Reconstruction                & Cos Sim$\uparrow$      & 0.798        & \yellow{0.849} & \yellow{0.891} & \yellow{0.920}  & \yellow{0.934}   & 0.610         & 0.826          & ---    \\ \hline
\multirow{2}{*}{Image Classification}  & Acc@1$\uparrow$        & 0.441        & 0.586        & \yellow{0.697} & \yellow{0.756}  & \yellow{0.776}   & 0.284         & 0.664          & \textcolor{gray}{0.841}  \\
                                       & Acc@5$\uparrow$        & 0.672        & 0.806        & \yellow{0.890} & \yellow{0.926}  & \yellow{0.938}   & 0.479         & 0.870          & \textcolor{gray}{0.969}  \\ \hline
\multirow{2}{*}{Semantic Segmentation} & mIoU$\uparrow$         & 0.250        & 0.372        & \yellow{0.480} & \yellow{0.536}  & \yellow{0.562}   & 0.059         & 0.410          & \textcolor{gray}{0.606}  \\ 
                                       & bIoU $\uparrow$        & 0.108        & 0.180        & \yellow{0.250} & \yellow{0.291}  & \yellow{0.309}   & 0.027         & 0.211          & \textcolor{gray}{0.343}  \\ \hline
\end{tabular}
}
\caption{\textbf{Evaluation of downstream tasks at different token lengths for One-D-Piece-S-256 and One-D-Piece-B-256}. \colorbox{green!10}{Green} background indicate where One-D-Piece-S-256 surpasses WebP, while \colorbox{yellow!20}{yellow} background show where One-D-Piece-B-256 surpasses WebP.}

\label{tab:downstream_tasks_sb256}
\end{table*}

\begin{table}[t!]
\setlength{\dashlinedash}{0.5pt}
\setlength{\dashlinegap}{1.5pt} 
\setlength{\arrayrulewidth}{0.5pt}
    \centering
    \begin{tabular}{ll}
        \hline
        \textbf{Item} & \textbf{Value} \\
        \hline \hline
        \multicolumn{2}{l}{\textbf{Model}} \\
        \hline
        Codebook Size & 4,096 \\
        Token Size & 12 \\
        Model Size & ViT small / base / large \\
        Patch Size & 16 \\
        Latent Tokens & 256 \\
        \hline
        \multicolumn{2}{l}{\textbf{Training}} \\
        \hline
        Stage1 Epochs & 100 \\
        Stage2 Epochs & 200 \\
        Stage1 Batch Size & 1024 \\
        Stage2 Batch Size & 512 \\
        Dataset & ImageNet-1K \\
        Augmentation & Random Crop / Flip \\
        \hline
        \multicolumn{2}{l}{\textbf{Losses}} \\
        \hline
        \multicolumn{2}{l}{\textbf{Stage1}} \\
        \hdashline
        Pretrained Tokenizer & MaskGIT tokenizer~\href{https://github.com/google-research/maskgit}{[Link]}\\
        Target Codebook Size & 1024 \\
        Reconstruction Weight & 1.0 \\
        Quantizer Weight & 1.0 \\
        \hdashline
        \multicolumn{2}{l}{\textbf{Stage2}} \\
        \hdashline
        Discriminator Weight & 0.01 \\
        Perceptual Loss Model & ConvNeXT-Small~\href{https://pytorch.org/vision/main/models/generated/torchvision.models.convnext_small.html#torchvision.models.ConvNeXt_Small_Weights}{[Link]}\\
        Perceptual Loss Weight & 0.1 \\
        Reconstruction Weight & 1.0 \\
        Commitment Loss Weight & 0.25 \\
        Codebook Loss Weight & 1.0 \\
        \hline
        \multicolumn{2}{l}{\textbf{Optimizer}} \\
        \hline
        Optimizer & AdamW \\
        Learning Rate & 1e-4 \\
        Beta1 & 0.9 \\
        Beta2 & 0.99 \\
        Weight Decay & 1e-4 \\
        Epsilon & 1e-8 \\
        \hline
        \multicolumn{2}{l}{\textbf{Scheduler}} \\
        \hline
        Scheduler Type & Cosine \\
        Warmup Steps & 10,000 \\
        End Learning Rate & 1e-5 \\
        \hline
    \end{tabular}
    \caption{
    \label{tab:hyperparameters}
    \textbf{Hyperparameters for One-D-Piece models}. These hyperparameters are fully following TiTok settings. }

\end{table}

\begin{table}[t!]
\setlength{\dashlinedash}{0.5pt}
\setlength{\dashlinegap}{1.5pt}
\setlength{\arrayrulewidth}{0.5pt}
    \centering
    \begin{tabular}{lccc}
        \hline
        \multirow{2}{*}{\textbf{Item}} & \multicolumn{3}{c}{\textbf{Value}} \\ \cmidrule{2-4}
        & \textbf{S-256} & \textbf{B-256} & \textbf{L-256} \\

        \hline \hline
        \multicolumn{4}{l}{\textbf{Model}} \\
        \hline
        Architecture & \multicolumn{3}{c}{MaskGIT} \\
        Hidden Dim & \multicolumn{3}{c}{768} \\
        Hidden Layers & \multicolumn{3}{c}{24} \\
        Attention Heads & \multicolumn{3}{c}{16} \\
        Dropout Rate & \multicolumn{3}{c}{0.1} \\
        Class Label Drop & \multicolumn{3}{c}{0.1} \\
        Class Count & \multicolumn{3}{c}{1000} \\
        Latent Tokens & \multicolumn{3}{c}{256} \\
        \hline
        \multicolumn{4}{l}{\textbf{Training}} \\
        \hline
        Epochs & \multicolumn{3}{c}{900} \\
        Batch Size & \multicolumn{3}{c}{2048} \\
        Dataset & \multicolumn{3}{c}{ImageNet-1K} \\
        Augmentation & \multicolumn{3}{c}{Random Flip} \\
        \hline
        \multicolumn{4}{l}{\textbf{Losses}}\\ 
        \hline
        Loss Function & \multicolumn{3}{c}{CrossEntropy} \\
        Label Smoothing & \multicolumn{3}{c}{0.1} \\
        Unmasked Token Loss & \multicolumn{3}{c}{0.1}\\
        \hline
        \multicolumn{4}{l}{\textbf{Optimizer}} \\
        \hline
        Optimizer     & \multicolumn{3}{c}{AdamW} \\
        Learning Rate & \multicolumn{3}{c}{2e-4} \\
        Beta1         & \multicolumn{3}{c}{0.9} \\
        Beta2         & \multicolumn{3}{c}{0.96} \\
        Weight Decay  & \multicolumn{3}{c}{0.03} \\
        \hline
        \multicolumn{4}{l}{\textbf{Scheduler}} \\
        \hline
        Scheduler Type    & \multicolumn{3}{c}{Cosine} \\
        Warmup Steps      & \multicolumn{3}{c}{10,000} \\
        End Learning Rate & \multicolumn{3}{c}{1e-5} \\
        \hline
        \multicolumn{4}{l}{\textbf{Decoding}} \\
        \hline
        Steps & 16 & 16 & 16 \\ 
        Temperature & 3.0 & 2.5 & 3.0 \\
        Guidance Decay & \multicolumn{3}{c}{Power Cosine~\tablefootnote{Shanghua Gao, Pan Zhou, Ming-Ming Cheng and Shuicheng Yan. Masked Diffusion Transformer is a Strong Image Synthesizer. in 2023 IEEE/CVF International Conference on Computer Vision (ICCV), Paris, France, 2023, pp. 23107-23116.}} \\
        Guidance Scale & 12.5 & 8.5 & 5.5   \\ \hline

    \end{tabular}
    \caption{
    \label{tab:generation_hyperparameters}
    \textbf{Hyperparameters for MaskGIT models}. These hyperparameters are fully following TiTok settings.}
\end{table}

\begin{figure*}[ht]
    \centering
    \includegraphics[width=0.80\linewidth]{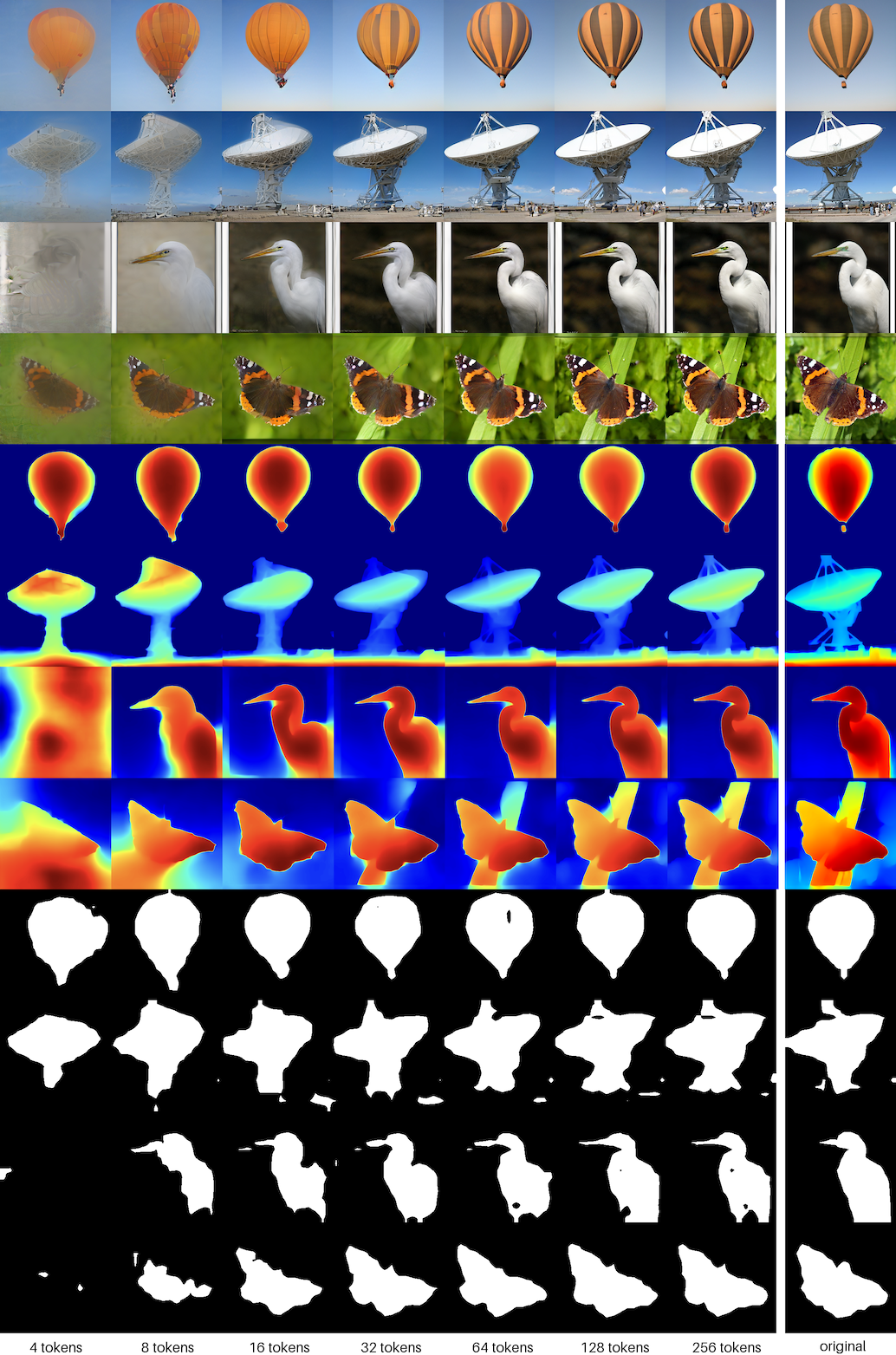}
    \caption{\textbf{Results of depth estimation and semantic segmentation on reconstructed images with One-D-Piece-L-256}. With an increase in token count, these results approach those of the original images.}
    \label{fig:depth_semantic_segmentation}
\end{figure*}

\subsection{Inference Speed}
\label{sec:speed}

While Tail Token Drop reduces the actual token count, it does not decrease the time complexity of the tokenizer. During tokenization, the inference process generates the maximum number of tokens, ensuring that the method introduces no difference on computational cost to the TiTok architecture. However, during detokenization, as fewer tokens are processed, some speed improvements can be observed.

\section{Analysis Details}

\begin{figure*}[htbp]
    \centering
    \includegraphics[width=0.9\linewidth]{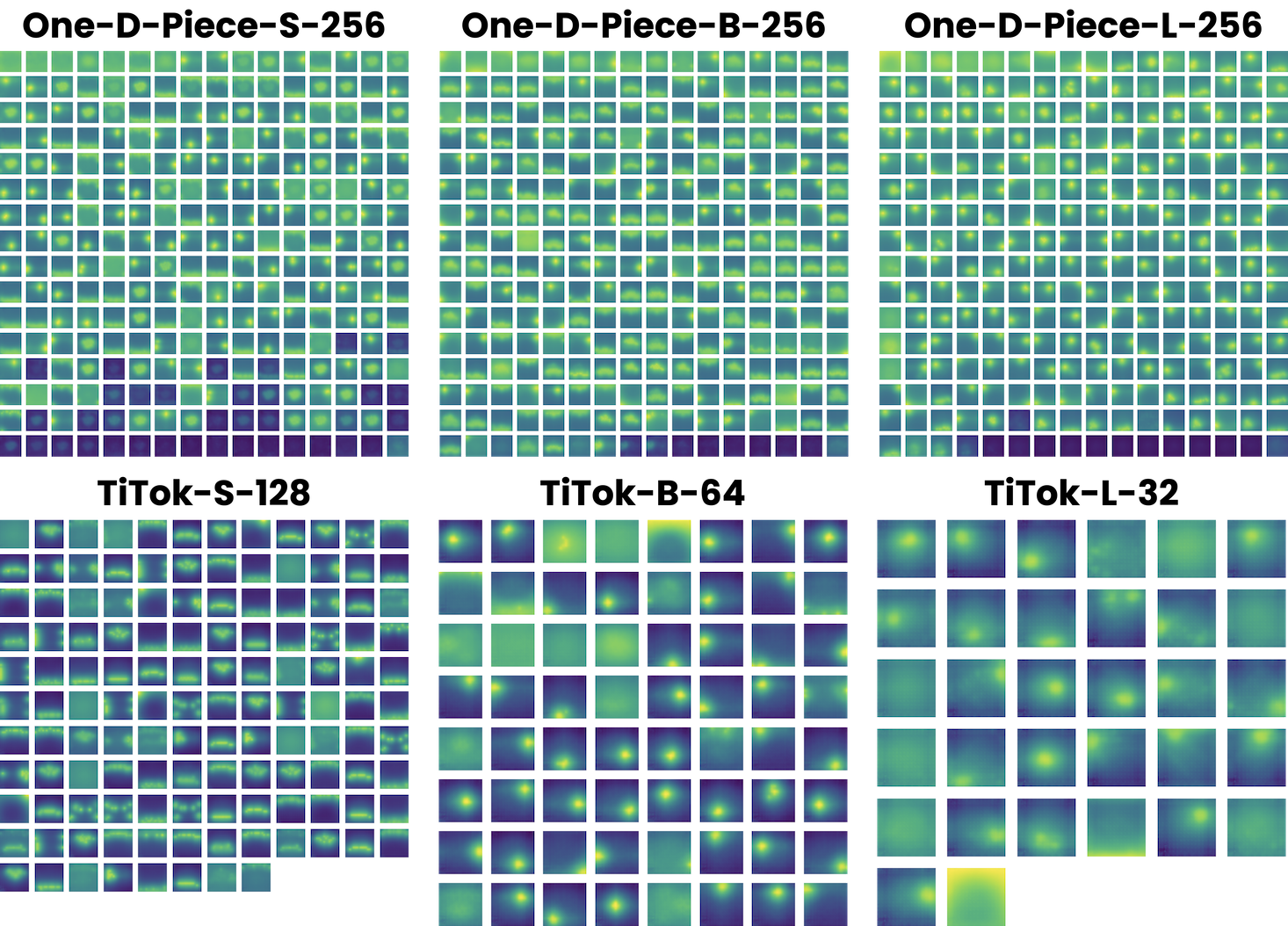}
    \caption{Token Contribution Grid for all variants of One-D-Piece and TiTok. Our One-D-Piece models demonstrate a clear concentration of global information at the head of the token sequence, whereas the TiTok models distribute such tokens more randomly.}
    \label{fig:token_contribution_grid_full}
\end{figure*}

We provide token contribution grids for all three One-D-Piece variants and three TiTok variants in \autoref{fig:token_contribution_grid_full}. Note that these visualizations are log-scaled and normalized using the global maximum and minimum values within each grid. As a result, the colors are not directly comparable across different models.

\section{Licenses}
The licenses of datasets and models used for training, evaluation, and downstream tasks are described as follows.

\paragraph{ImageNet-1K} 
ImageNet-1K comprises 1,281,167 training images, 50,000 validation images, and 100,000 test images, covering a total of 1,000 object classes. We use it under \href{https://image-net.org/accessagreement}{the terms of its access agreement}, which permits usage for non-commercial research and educational purposes.

\paragraph{ImageNet-S}
ImageNet-S provides high-quality semantic segmentation annotations for robust evaluation, based on 12,419 validation images and 26,423 test images sourced from ImageNet. The dataset focuses on 919 categories, excluding unsegmentable ones such as ``bookshop.'' Usage of this dataset adheres to the \noindent \href{https://image-net.org/accessagreement}{ImageNet licensing terms}.

\paragraph{COCO}
COCO val2017 consists of 5,000 images spanning 80 categories, including a wide range of annotated objects such as people, animals, vehicles, and furniture. We follow \href{https://cocodataset.org/#termsofuse}{the terms of use of COCO} and \noindent \href{https://www.flickr.com/help/terms}{the Flickr Terms of Use} for images in the COCO dataset.

\paragraph{Ultralytics YOLO11}
We use Ultralytics YOLO11x for object detection. Ultralytics YOLO11 is released under \href{https://github.com/ultralytics/ultralytics/blob/main/LICENSE}{the GNU Affero General Public License v3.0 (AGPL-3.0)}, which permits the use of the model for research purposes. 

\end{document}